\documentclass[a4paper, 10pt]{article}
\usepackage{cite}
\usepackage[pdftex]{graphicx}
\usepackage[left=1.5cm, right=1.5cm, bottom=2cm, top=2cm]{geometry}
\usepackage{amsmath}
\usepackage{amsfonts}
\usepackage{amssymb}
\usepackage{algorithm}
\usepackage{algorithmicx}
\usepackage{algpseudocode}
\usepackage{array}
\usepackage{fixltx2e}
\usepackage{url}
\usepackage{multirow}
\usepackage{float}
\usepackage{lscape}
\usepackage[caption = false]{subfig}
\usepackage{balance}
\usepackage{lscape}

\begin{document}
%
\title{Generative Adversarial Networks for Financial \\ Trading Strategies Fine-Tuning and Combination}

\author{Adriano~Koshiyama, Nick~Firoozye, and~Philip~Treleaven \\ 
Department
of Computer Science, University College London \\ Gower Street, London WC1E 6BT Tel: +44 (0) 20 7679 2000, United Kingdom \\ 
\url{[adriano.koshiyama.15, n.firoozye, p.treleaven]@ucl.ac.uk}}

%
%

\maketitle

\begin{abstract}
Systematic trading strategies are algorithmic procedures that allocate assets aiming to optimize a certain performance criterion. To obtain an edge in a highly competitive environment, the analyst needs to proper fine-tune its strategy, or discover how to combine weak signals in novel alpha creating manners. Both aspects, namely fine-tuning and combination, have been extensively researched using several methods, but emerging techniques such as Generative Adversarial Networks can have an impact into such aspects. Therefore, our work proposes the use of Conditional Generative Adversarial Networks (cGANs) for trading strategies calibration and aggregation. To this purpose, we provide a full methodology on: (i) the training and selection of a cGAN for time series data; (ii) how each sample is used for strategies calibration; and (iii) how all generated samples can be used for ensemble modelling. To provide evidence that our approach is well grounded, we have designed an experiment with multiple trading strategies, encompassing 579 assets. We compared cGAN with an ensemble scheme and model validation methods, both suited for time series. Our results suggest that cGANs are a suitable alternative for strategies calibration and combination, providing outperformance when the traditional techniques fail to generate any alpha.

\end{abstract}

\twocolumn

\section{Introduction}
Systematic trading strategies are algorithmic procedures that allocate assets aiming to optimize a certain performance criterion. To obtain an edge in a highly competitive environment, the analyst needs to proper fine-tune its strategy, or discover how to combine weak signals in novel alpha creating manners. Both aspects, fine-tuning and combination, have been extensively researched in different domains, with distinct emphasis and assumptions:

\begin{itemize}
	\item Forecasting and Financial Econometrics: proper model fine-tuning is also known as preventing {\em Backtesting Overfitting}: partly due to the endemic abuse of backtested results, there is an increasing interest in devising procedures for the assessment and comparison of strategies \cite{harvey2015backtesting,romano2016efficient,bailey2015probability}. {\em Model/Forecasting combination} is an established area of research \cite{timmermann2006forecast}, starting with the seminar work of \cite{bates1969combination} in the 60s, and still active \cite{hsiao2014there}.
	
	\item Computational Statistics and Machine Learning: model tuning falls under the guise of {\em Hyperparameter Optimization} \cite{eggensperger2013towards,loshchilov2016cma} and {\em Model Validation} schemes \cite{arlot2010survey,lahiri2013resampling}; research on their interaction is scarce, and dealing with dependent data scenarios is still an open area of research \cite{jiang2017markov,bergmeir2018note}. Forming ensembles is a modelling strategy widely adopted by this community, being Random Forest and Gradient Boosting Trees the two main workhorses of Bagging and Boosting strategies \cite{friedman2001elements,efron2016computer}. 
\end{itemize}

In summary, proper model tuning and combination are still an active area of research, in particular to dependent data scenarios (e.g., time series). Emerging techniques such as Conditional Generative Adversarial Networks \cite{mirza2014conditional} can have an impact into aspects of trading strategies, specifically fine-tuning and to form ensembles. Also, we can list a few advantages of such method, like: (i) generating more diverse training and testing sets, compared to traditional resampling techniques; (ii) able to draw samples specifically about stressful events, ideal for model checking and stress testing; and (iii) providing a level of anonymization to the dataset, differently from other techniques that (re)shuffle/resample data.

The price paid is having to fit this generative model for a given time series. In this work we show how the training and selection of the generator is made; overall, this part tends to be less costly than the whole backtesting or ensemble modelling process. Therefore, our work proposes the use of Conditional Generative Adversarial Networks (cGANs) for trading strategies calibration and aggregation. We provide evidence that cGANs can be used as tools for model fine-tuning, as well as on setting up  ensemble of strategies. Hence, we can summarize the main highlights of this work:

\begin{itemize}
	\item We have considered 579 assets, mainly equities, but we have also included swaps and equity indices, and currencies data.
	\item Our findings suggest that cGAN can be an alternative to Bagging via Stationary Bootstrap, that is, when bootstrap approaches have failed to outperform, cGAN can be employed for Stochastic Gradient Boosting or Random Forest. 
	\item For model fine-tuning, we have evidence that cGAN is a viable procedure, comparable to many other well established techniques. Therefore, it should be considered part of the quantitative strategist toolkit of validation schemes for time series modelling.
	\item A side outcome of our work is the wealth of results and comparisons: to the best of our knowledge most of the applied model validation strategies have not yet been cross compared using real datasets and different models.
\end{itemize}

Therefore, in addition to this introductory section, we structured this paper with other four sections. Next section provides background information about GANs and cGANs, how the training and selection of cGANs for time series was performed, as well as their application to fine-tuning and ensemble modelling of trading strategies. Third section outlines the experimental setting (scenario, parameters, etc.) used for two case studies: fine-tuning and combination of trading strategies. After this, section IV presents the results and discussion of both case studies, with section V exhibiting our concluding remarks and potential future works.

\section{Generative Adversarial Networks}

\subsection{Background}

Generative Adversarial Networks (GANs) \cite{goodfellow2014generative} is a modelling strategy that employ two Neural Networks: a Generator ($G$) and a Discriminator ($D$). The Generator is responsible to produce a rich, high dimensional vector attempting to replicate a given data generation process; the Discriminator acts to separate the input created by the Generator and of the real/observed data generation process. They are trained jointly, with $G$ benefiting from $D$ incapability to recognise true from generated data, whilst $D$ loss is minimized when it is able to classify correctly inputs coming from $G$ as fake and the dataset as true. Competition drive both Networks to improve their performance until the genuine data is indistinguishable from the generated one.

From a mathematical perspective, we start by defining a prior $p_{\mathbf{z}}(\mathbf{z})$ on input noise variables $\mathbf{z}$, which will be used by the Generator, denoted as a neural network $G(\mathbf{z}, \Theta_G)$ with parameters $\Theta_G$, that maps noise to the data/input space $G: \mathbf{z} \rightarrow \mathbf{x}$. We also need to set the Discriminator, represented as a neural network $D(\mathbf{x}^\ast, \Theta_D)$, that scores how likely is that $\mathbf{x}^\ast$ was sampled from the dataset ($p_{data}(\mathbf{x})$ -- $D: \mathbf{x}^\ast \rightarrow [0, 1]$). As outlined before, $D$ is trained to maximize correct labelling, whilst $G$, in the original formulation, is trained to minimize $\log(1- D(G(\mathbf{z})))$. It follows from \cite{goodfellow2014generative} that $D$ and $G$ play the following two-player minimax game with value function $V(G,D)$:
\begin{eqnarray}
\min_G \max_D V(D,G) = \mathbb{E}_{\mathbf{x} \sim p_{data}(\mathbf{x})} [\log D(\mathbf{x})] + \nonumber \\ \mathbb{E}_{\mathbf{z} \sim p_{\mathbf{z}}(\mathbf{z})} [\log (1 - D(G(\mathbf{z})))]
\end{eqnarray}

Overall, GANs have been successfully applied to image and text generation \cite{creswell2018generative:paper2}. However, some issues linked to its training and applications to special cases \cite{salimans2016improved,gulrajani2017improved} have fostered a substantial amount of research in newer architectures, loss functions, training, etc. We can classify and summarise these new methods as of belonging to: 
\begin{itemize}
	\item Ensemble Strategies: train multiple GANs, with different initial conditions, slices of data and tasks to attain; orchestrate the generator output by employing an aggregation operator (summing, weighted averaging, etc.), from multiple checkpoints or at the end of the training. Notorious instantiations of these steps are the Stacked GANs \cite{huang2017stacked}, Ensembles of GANs \cite{wang2016ensembles} and AdaGANs \cite{tolstikhin2017adagan}.
	\item Loss Function Reshaping: reshape the original loss function (a lower-bound on the Jensen-Shannon divergence) so that issues linked to training instability can be circumvented. Typical examples are: employing Wasserstein-1 Distance with a Gradient Penalty \cite{arjovsky2017wasserstein,gulrajani2017improved}; using a quantile regression loss function to implicitly push $G$ to learn the inverse of a cumulative density function \cite{ostrovski2018autoregressive}; rewriting the objective function with a mean squared error form -- now minimizing the $\chi^2$-distance \cite{mao2017least}; or even view the discriminator as an energy function that assigned low energy values to the regions of high data density, guiding the generator to sample from those regions \cite{zhao2016energy}.
	\item Adjusting Architecture and Training Process: we can mention Deep Convolutional GAN \cite{radford2015unsupervised}, in which a set of constraints on the architectural topology of Convolutional GANs are put in place to make the training more stable. Also, Evolutionary GANs \cite{wang2018evolutionary} that adds to the training loop of a GAN different metrics to jointly optimize the generator, as well as employing a population of Generators, created by introducing novel mutation operators.
\end{itemize}

Another issue, more associated to our work, is the handling of time series since learning an unconditional model, similar to the original formulation, works for image and text creation/discovery. However, when the goal is to use it for time series modelling, a sampling process that can take into account the previous state space is required to preserve the time series statistical properties (autocorrelation structure, trends, seasonality, etc.). In this sense, next subsection deals with Conditional GANs \cite{mirza2014conditional}, a more appropriate modelling strategy to handle dependent data generation.

\subsection{Conditional GANs}

As the name implies, Conditional GANs (cGANs) are an extension of a traditional GAN, when both $G$ and $D$ decision is based not only in noise or generated inputs, but include an additional information set $\mathbf{v}$. For example, $\mathbf{v}$ can represent a class label, a certain categorical feature, or even a current/expected market condition; hence, cGAN attempts to learn an implicit conditional generative model. Such application is more appropriate in cases where the data follows a sequence (time series, text, etc.) or when the user wants to build "what-if" scenarios (given that S\&P 500 has fallen 1\%, how much should I expect in basis points change of a US 10 year treasury?).

Most of the applications of cGANs related to our work have centred in synthesizing data to improve supervised learning models. The only exception is \cite{zhou2018stock}, where the authors use a cGAN to perform direction prediction in stock markets. Works \cite{fiore2017using,douzas2018effective} deal with the problem of imbalanced classification, in particular to fraud detection; they are able to show that cGANs compare favourably to other traditional techniques for oversampling. In \cite{esteban2017real}, the one that is closest to our work, the authors propose to use cGANs for medical time series generation and anonymization. They used cGANs to generate realistic synthetic medical data, so that this data could be shared and published without privacy concerns, or even used to augment or enrich similar datasets collected in different or smaller cohorts
of patients. 

\begin{algorithm*}[h!]
	\caption{cGAN Training and Selection}\label{training}
	\begin{algorithmic}[1]
		\Procedure{cGAN}{$[y_1,...,y_T], params$}
		\For{number of epochs}
		\State Sample minibatch of $L$ noise samples $\{\mathbf{z}^{(1)}, ..., \mathbf{z}^{(L)}\}$ from noise prior $p_{\mathbf{z}} (\mathbf{z})$
		\State Sample minibatch of $L$ examples $\{(y_t; y_{t-1}, ..., y_{t-p})^{(1)}, ..., (y_t; y_{t-1}, ..., y_{t-p})^{(L)}\}$ from $p_{data}(y_t | y_{t-1}, ..., y_{t-p})$
		\State Update the discriminator by ascending its stochastic gradient:
		\begin{equation}
		\nabla_{\Theta_D} \frac{1}{L} \sum_{l=1}^{L} \Big[\log D(y_t^{(l)} | y_{t-1}^{(l)}, ..., y_{t-p}^{(l)}) + \log (1 - D(G(\mathbf{z}^{(l)} | y_{t-1}^{(l)}, ..., y_{t-p}^{(l)})))\Big] \nonumber
		\end{equation}
		\State Sample minibatch of $L$ noise samples $\{\mathbf{z}^{(1)}, ..., \mathbf{z}^{(L)}\}$ from noise prior $p_{\mathbf{z}} (\mathbf{z})$
		\State Update the generator by ascending its stochastic gradient:
		\begin{equation}
		\nabla_{\Theta_G} \frac{1}{L} \sum_{l=1}^{L} \Big[\log (D(G(\mathbf{z}^{(l)} | y_{t-1}^{(l)}, ..., y_{t-p}^{(l)})))\Big] \nonumber
		\end{equation}
		\If{$rem(epoch, snap) == 0$}
		\State $G_k \gets G$, $D_k \gets D$ \Comment{store current $G$, $D$ as $G_k$, $D_k$}
		\For{$c \gets 1, C$} \Comment{draw $C$ samples from $G_k$}
		\For{$t\gets p+1, T$} \Comment{generate time series}
		\State sample noise vector $\mathbf{z} \sim p_{\mathbf{z}} (\mathbf{z})$
		\State draw $y_t^\ast = G_k(\mathbf{z} | y_{t-1}...., y_{t-p})$
		\EndFor
		\State Measure cGAN sample goodness-of-fit (akin to chi-square distance): 
		\begin{equation}
		RMSE_c = \sqrt{\frac{1}{T-p-1} \sum_{p+1}^{T} (y_t - y_t^\ast)^2} \nonumber
		\end{equation}
		\EndFor
		\State Average of all samples: $RMSE(G_k) = \frac{1}{C} \sum_{c=1}^{C} RMSE_c$
		\EndIf
		\EndFor
		\State \textbf{return} $G := arg\min_{G_k} RMSE(G_k)$, $D := arg\min_{G_k} RMSE(G_k)$
		\EndProcedure
	\end{algorithmic}
\end{algorithm*}

Formally, we can define a cGAN by including the conditional variable $\mathbf{v}$ in the original formulation. Therefore, now $G: \mathbf{z} \times \mathbf{v} \rightarrow \mathbf{x}$ and $D: \mathbf{x}^\ast \times \mathbf{v} \rightarrow [0, 1]$, as before $D$ is trained to maximize correct labelling, whilst $G$, in the original formulation, is trained to minimize $\log(1- D(G(\mathbf{z} | \mathbf{v})))$. Similarly, it follows from \cite{mirza2014conditional} that $D$ and $G$ play the following two-player minimax game with value function $V(G,D)$:
\begin{eqnarray}
\min_G \max_D V(D,G) = \mathbb{E}_{\mathbf{x} \sim p_{data}(\mathbf{x})} [\log D(\mathbf{x} | \mathbf{v})] + \nonumber \\ \mathbb{E}_{\mathbf{z} \sim p_{\mathbf{z}}(\mathbf{z})} [\log (1 - D(G(\mathbf{z} | \mathbf{v})))]
\end{eqnarray}

\noindent in our case, given a time series $y_1, y_2, ..., y_t, ..., y_T$, our conditional set is $\mathbf{v} = (y_{t-1}, y_{t-2}, ..., y_{t-p})$ and what we are aiming to sample/discriminate is $\mathbf{x} = y_{t}$ (with $p_{data}$ $(y_t | y_{t-1}, ..., y_{t-p})$). In this sense, $p$ sets the amount of past information that is considered in the implicit conditional generative model. If $p = 0$, then a traditional GAN will be trained; if $p$ is large, than the Neural Network have a larger memory, but it will need bigger capacity to model and deal with selecting the right past values and dealing with noise vector $\mathbf{z}$; experimental setting section outline the values we have used during our experiments.

\subsection{Training and Selecting Generators for Time Series}

With the addition of the conditional vector $\mathbf{v}$, training cGANs is akin to GANs; what substantially change is how the right architecture is chosen across the training. Algorithm \ref{training} detail a minibatch stochastic gradient descent Training and Selecting of cGANs. 

\vspace{0.25cm}

\noindent $params$ represents a set of hyperparameters that the user has to define before running \textit{cGAN Training}. It mainly encompasses: $G$ and $D$ architectures, number of lags $p$, noise vector size and prior distribution, minibatch size $L$, number of epochs, snapshot frequency ($snap$), number of samples $C$, and parameters associated to the stochastic gradient optimizer; all of them are specified in the Experimental Setting section (see Table \ref{cgan-configs}).

\begin{figure*}[h!]
	\centering
	\subfloat[]{\includegraphics[width=.33\linewidth]{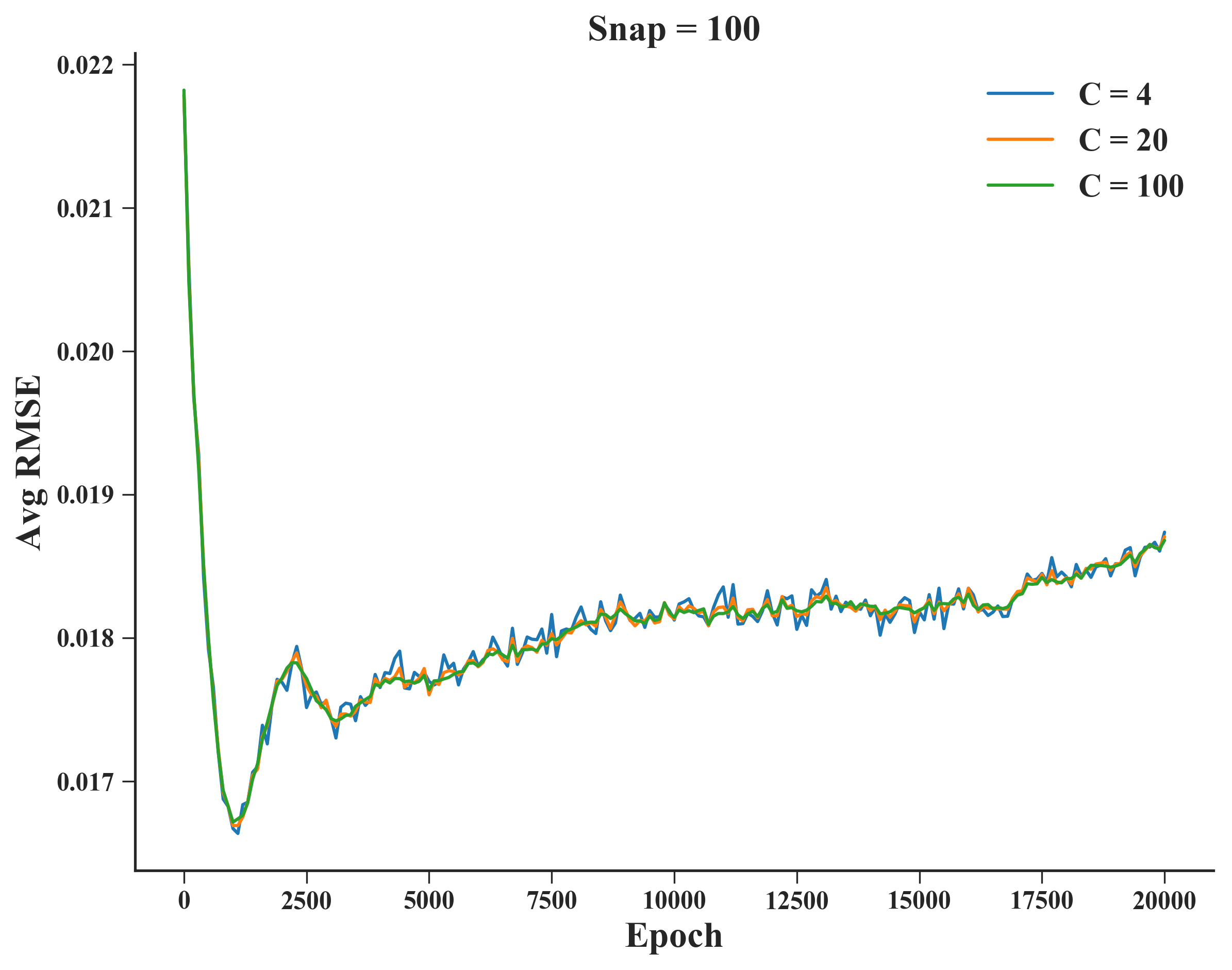}} \subfloat[]{\includegraphics[width=.33\linewidth]{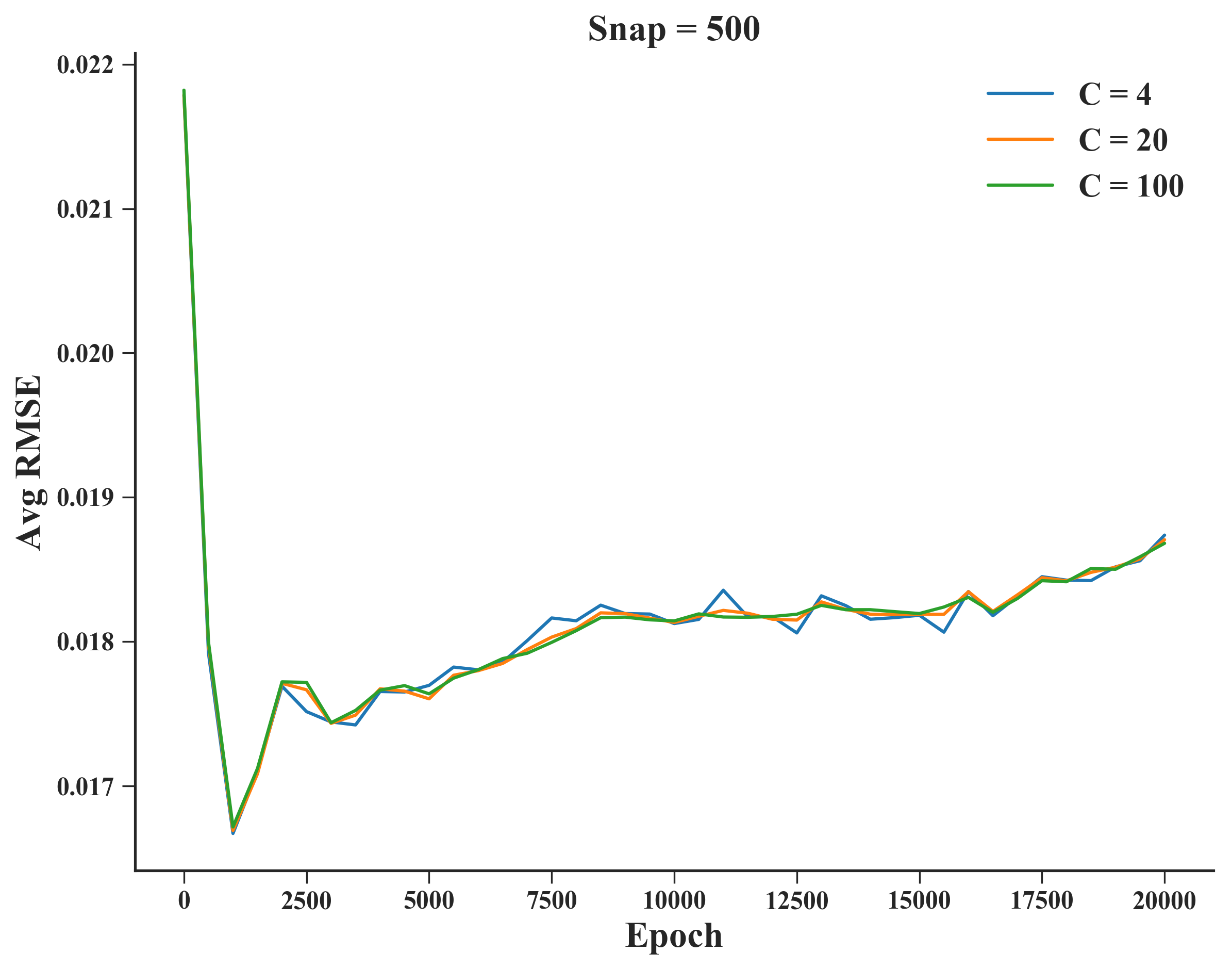}}
	\subfloat[]{\includegraphics[width=.33\linewidth]{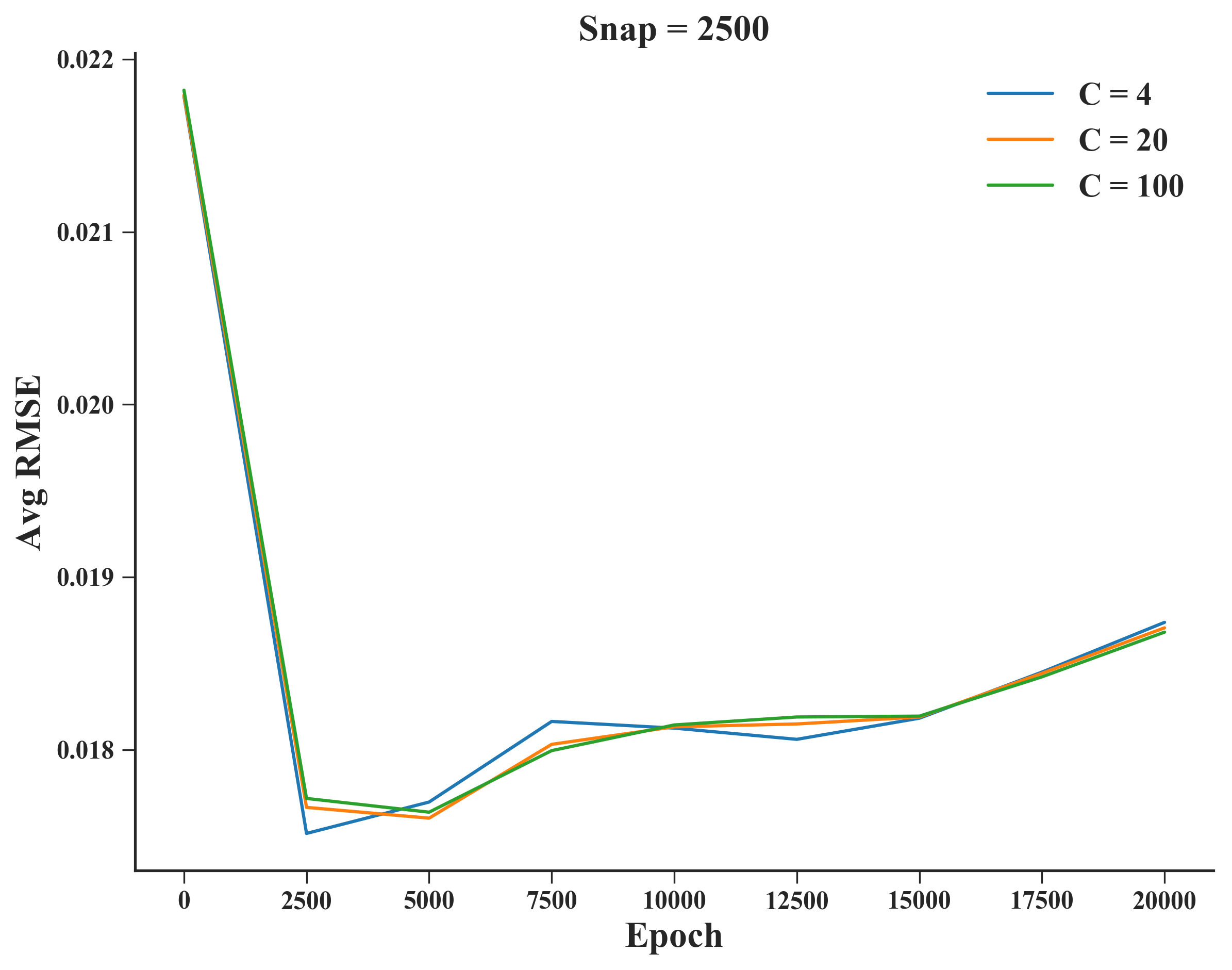}} 
	\caption{RMSE curves, considering a range of snapshot frequencies and number of samples.}
	\label{spx-cgan-rmse}
\end{figure*}

\begin{figure*}[h!]
	\centering
	\subfloat[]{\includegraphics[width=.33\linewidth]{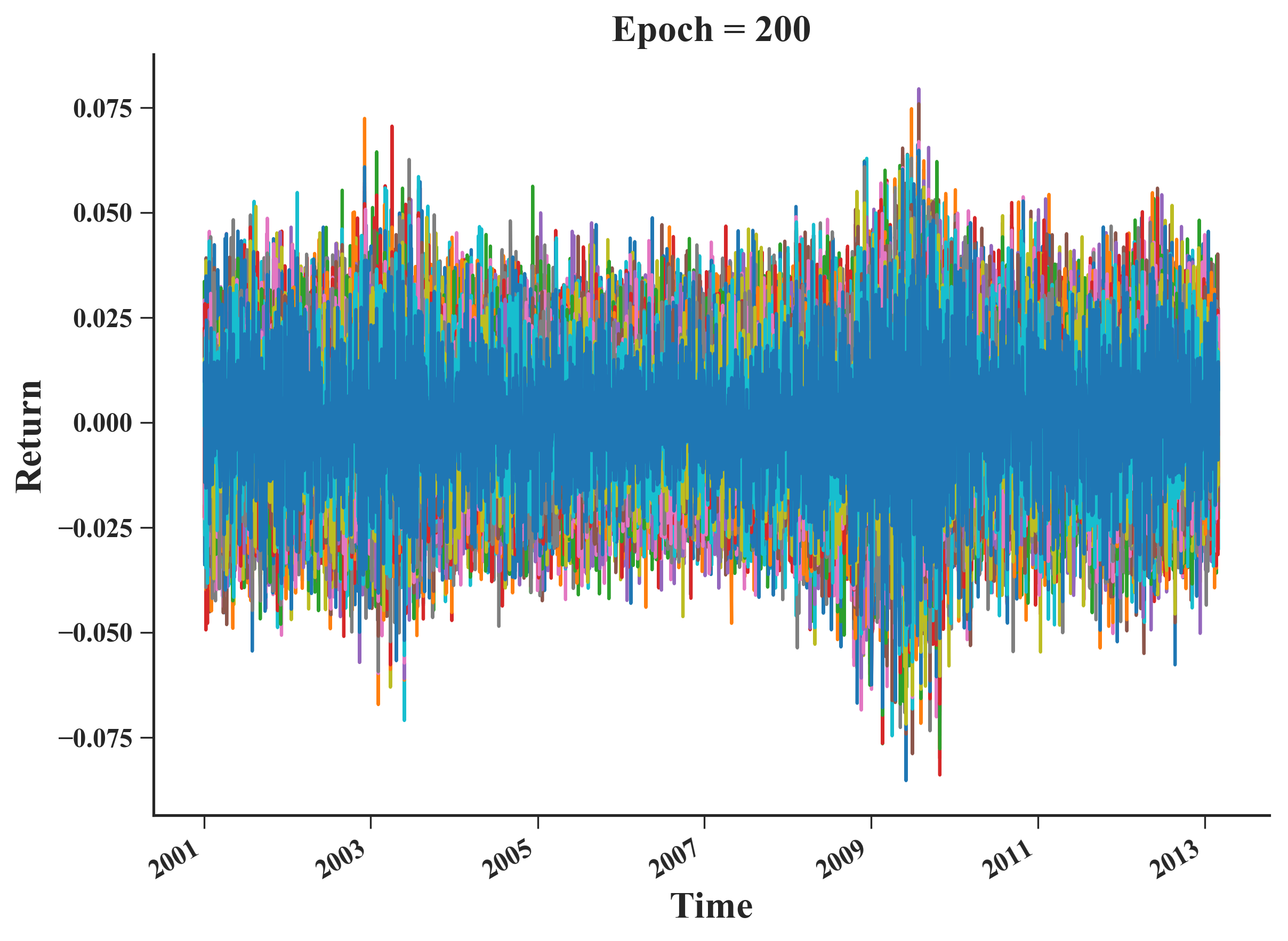}} \subfloat[]{\includegraphics[width=.33\linewidth]{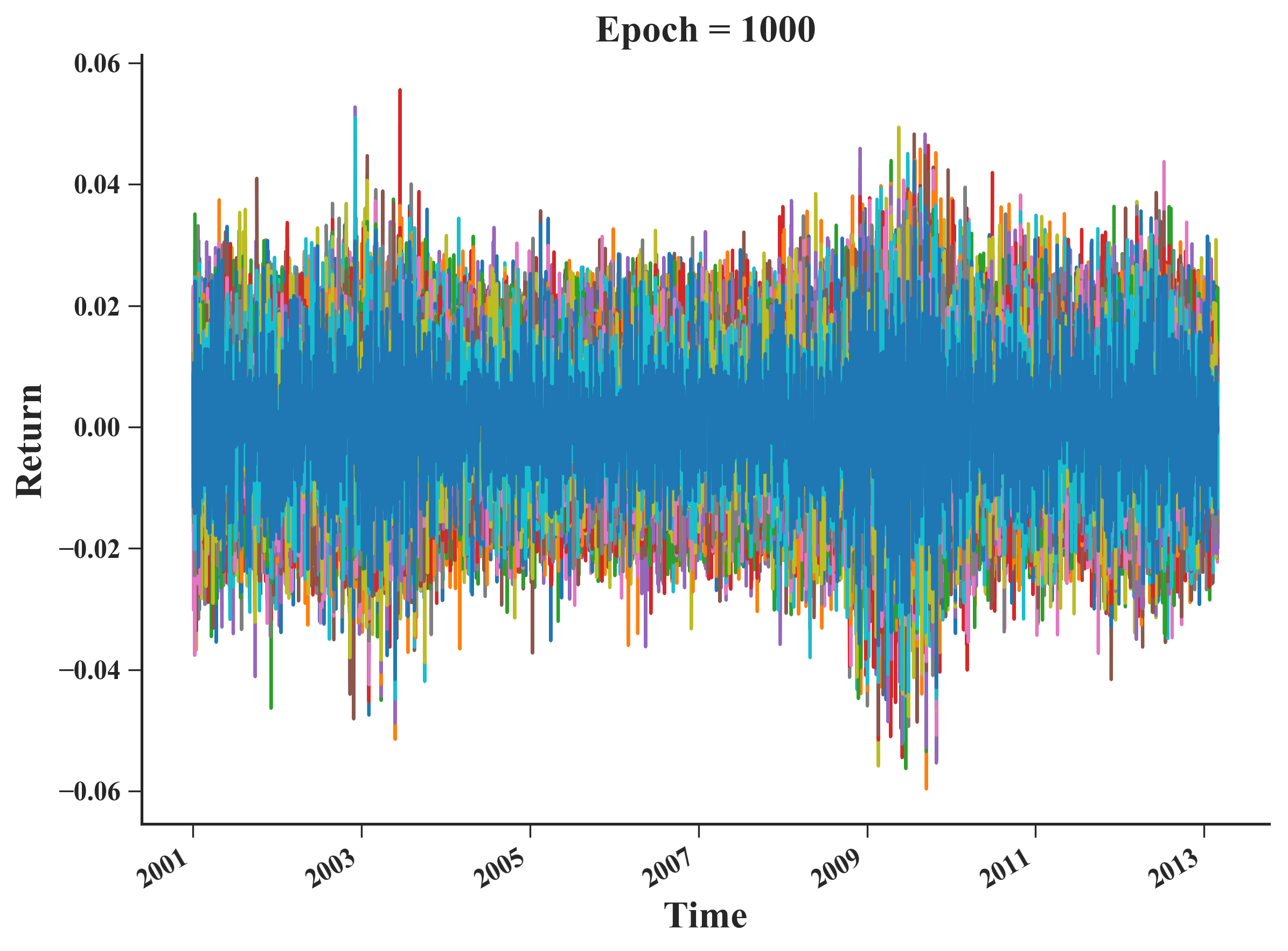}}
	\subfloat[]{\includegraphics[width=.33\linewidth]{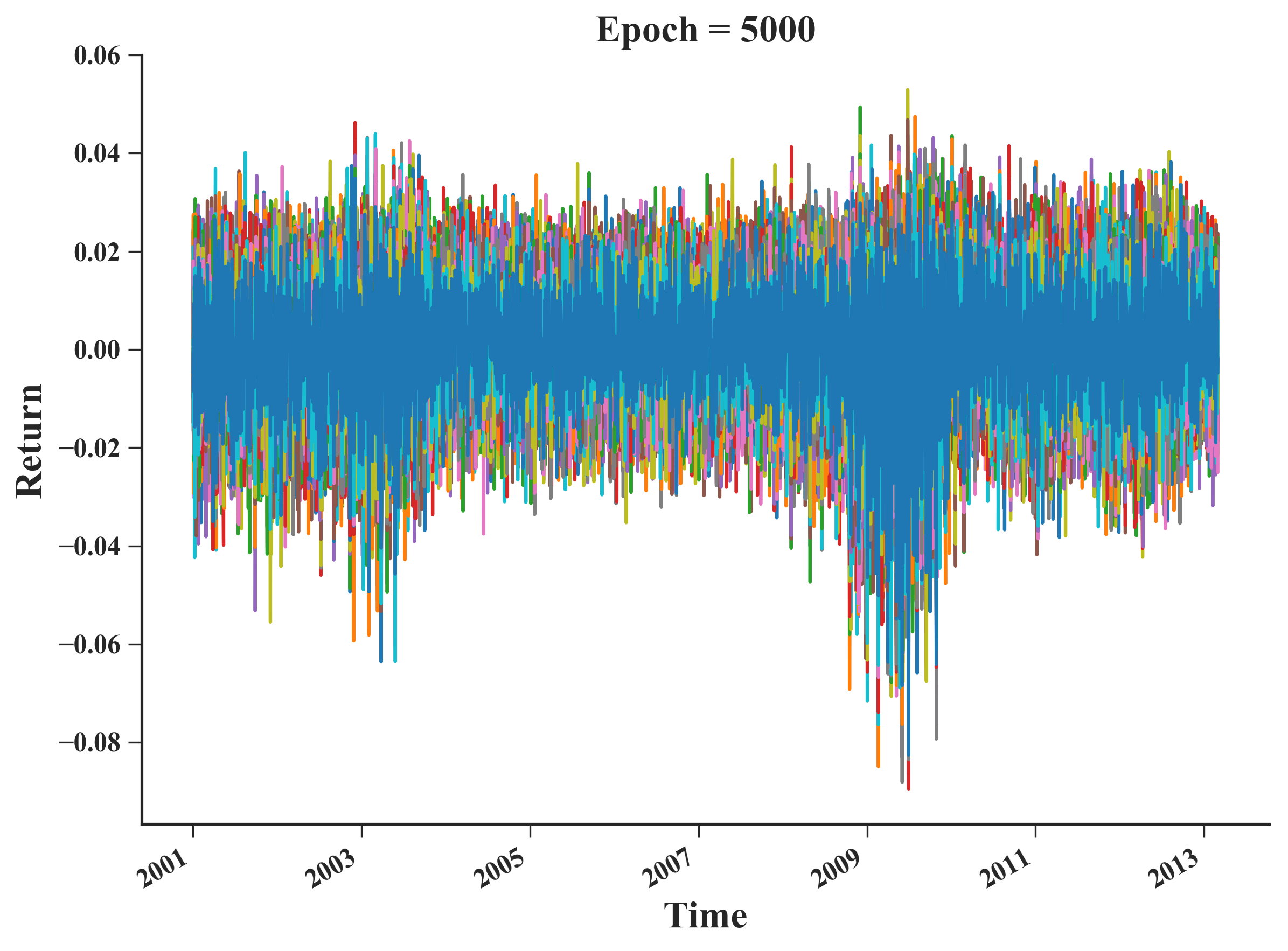}} \\
	\subfloat[]{\includegraphics[width=.33\linewidth]{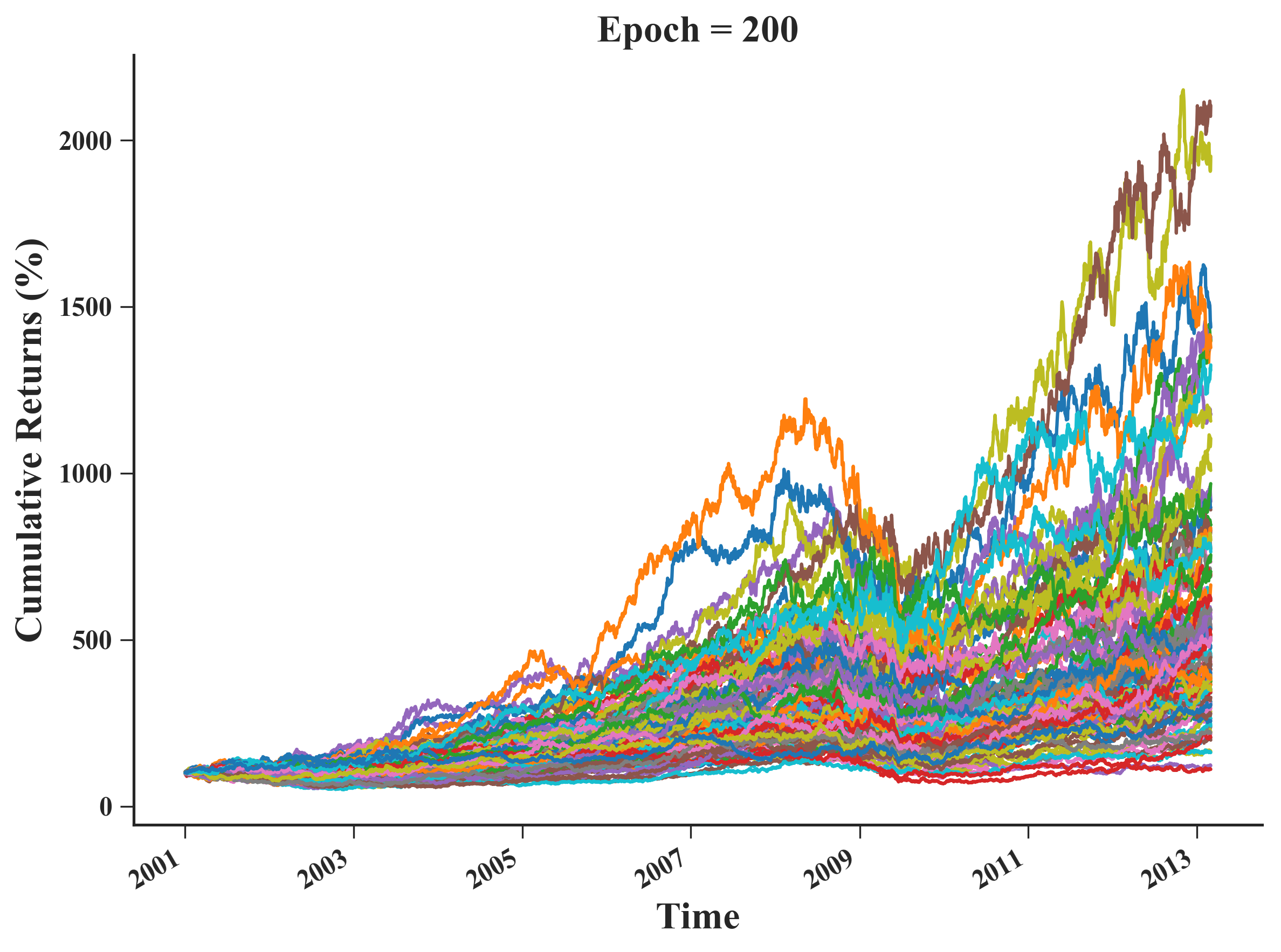}} \subfloat[]{\includegraphics[width=.33\linewidth]{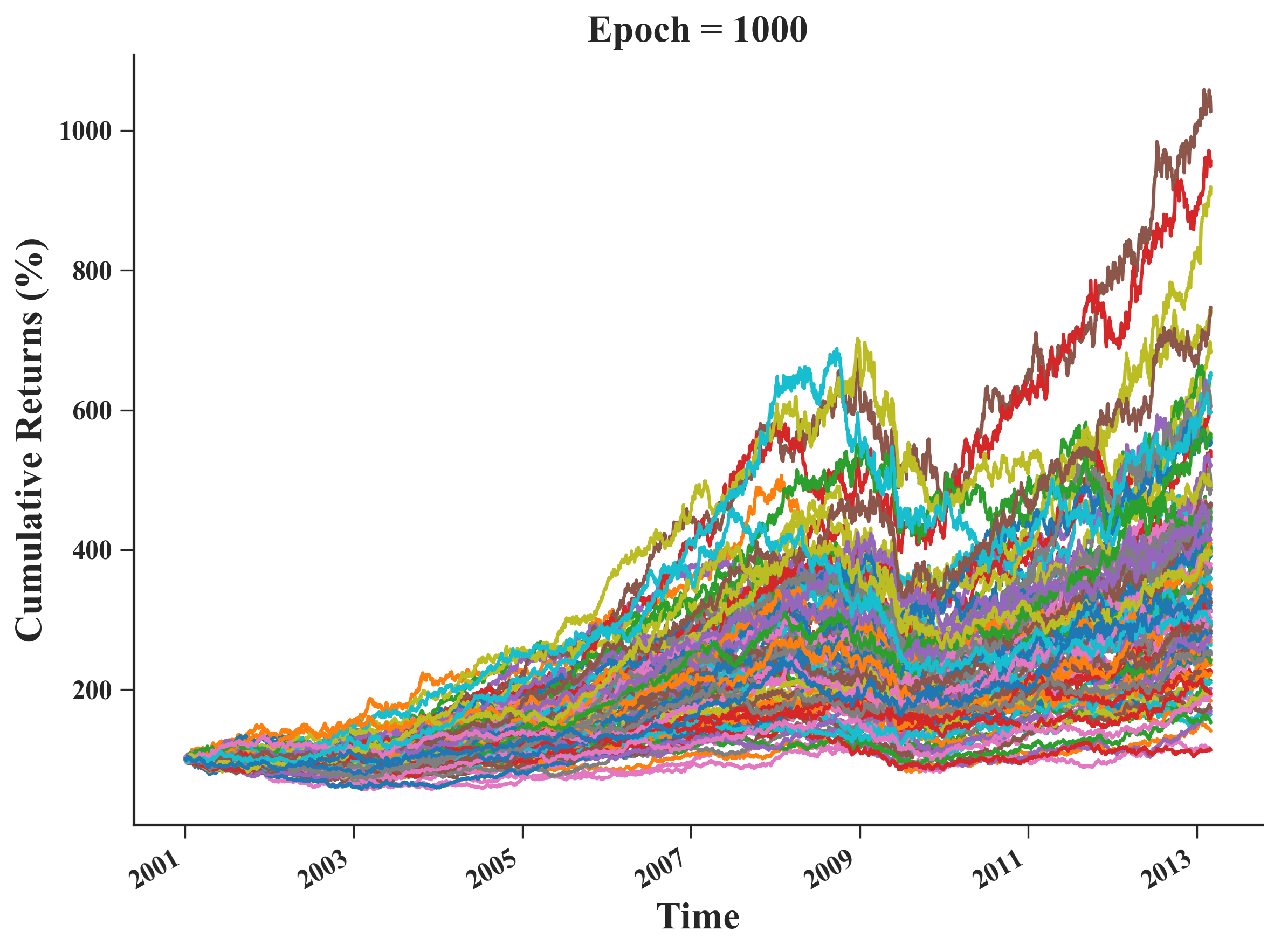}}
	\subfloat[]{\includegraphics[width=.33\linewidth]{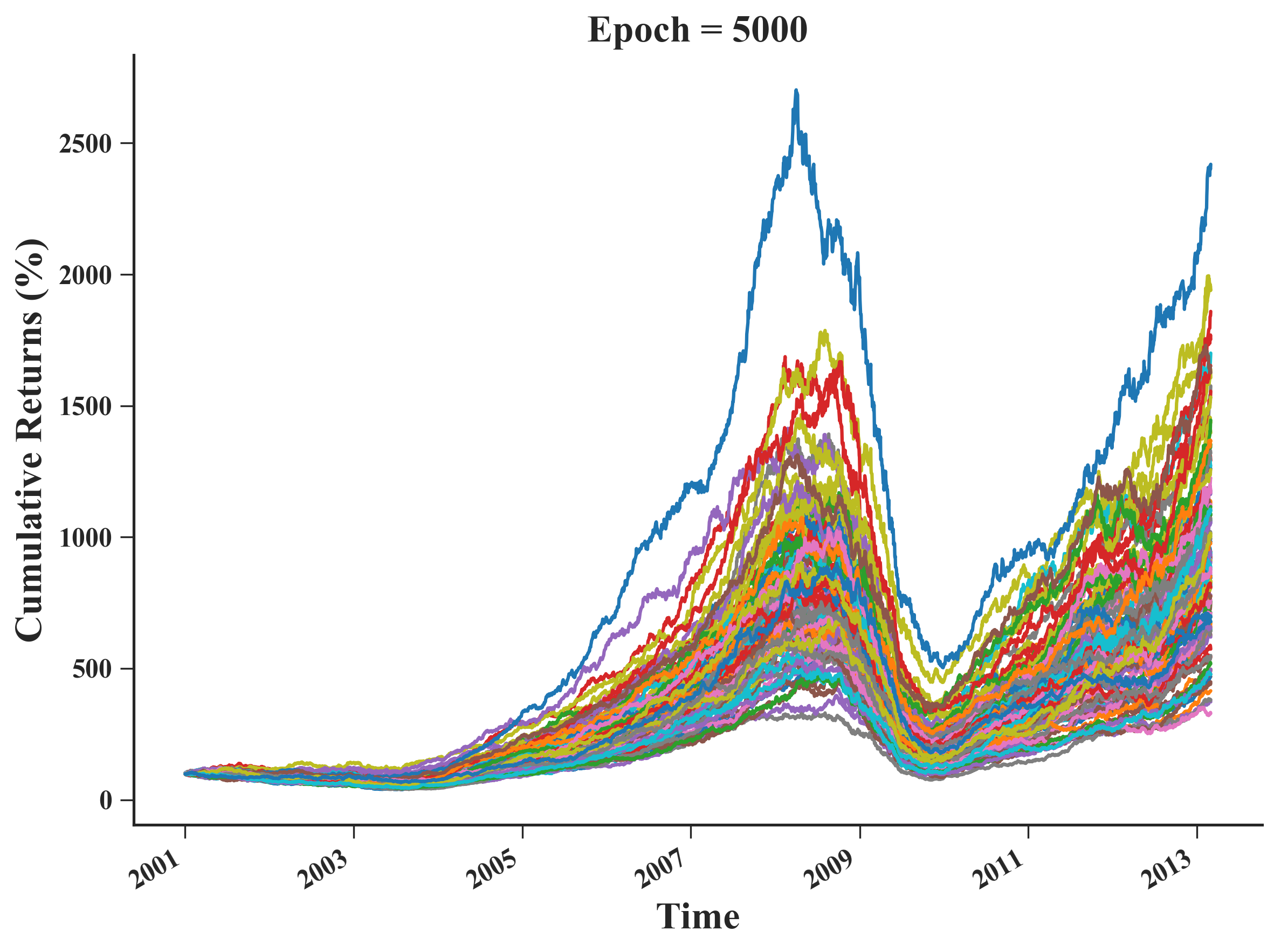}}
	\caption{SPX Index log-returns (a-c) from cGAN in different epochs, with their respective cumulative returns (d-e).}
	\label{spx-cgan-cumreturns}
\end{figure*}

Selecting the right cGAN during the training is a difficult task, since it is computationally expensive to every iteration draw multiple samples and evaluate them. An approximation that we considered was to add a snapshot frequency in which every $snap$ iterations $G$ and $D$ weights are store. This parameter plays a relevant role in regulating the available number of cGANs to draw samples, evaluate and select. To illustrate the selection part of Algorithm \ref{training}, Figure \ref{spx-cgan-rmse} presents a sensibility analysis of it to SPX Index. 

Overall, after a sharp decrease in the first 2000 epochs we observe a stabilization of RMSE to the 0.018 level. Drawing more samples improve estimation, but the gain is almost imperceptible from 20 to 100 samples. Snapshot frequency is an important parameter, with a noticeable difference between 100 to 2500, but without much change moving from 100 to 500. Number of samples draw from $G$ and the snapshot frequency are also reported in the Experimental Setting section. Figure \ref{spx-cgan-cumreturns} presents SPX Index samples (a-c) from cGAN and their respective cumulative returns (d-e)\footnote{We are highlighting this period in particular because our analyses and results concentrated on taking samples from 2001-2013.} in different stages of training: 200, 1000 and 5000 epochs. 

\begin{figure}[h!]
	\centering
	\subfloat[]{\includegraphics[width=\columnwidth]{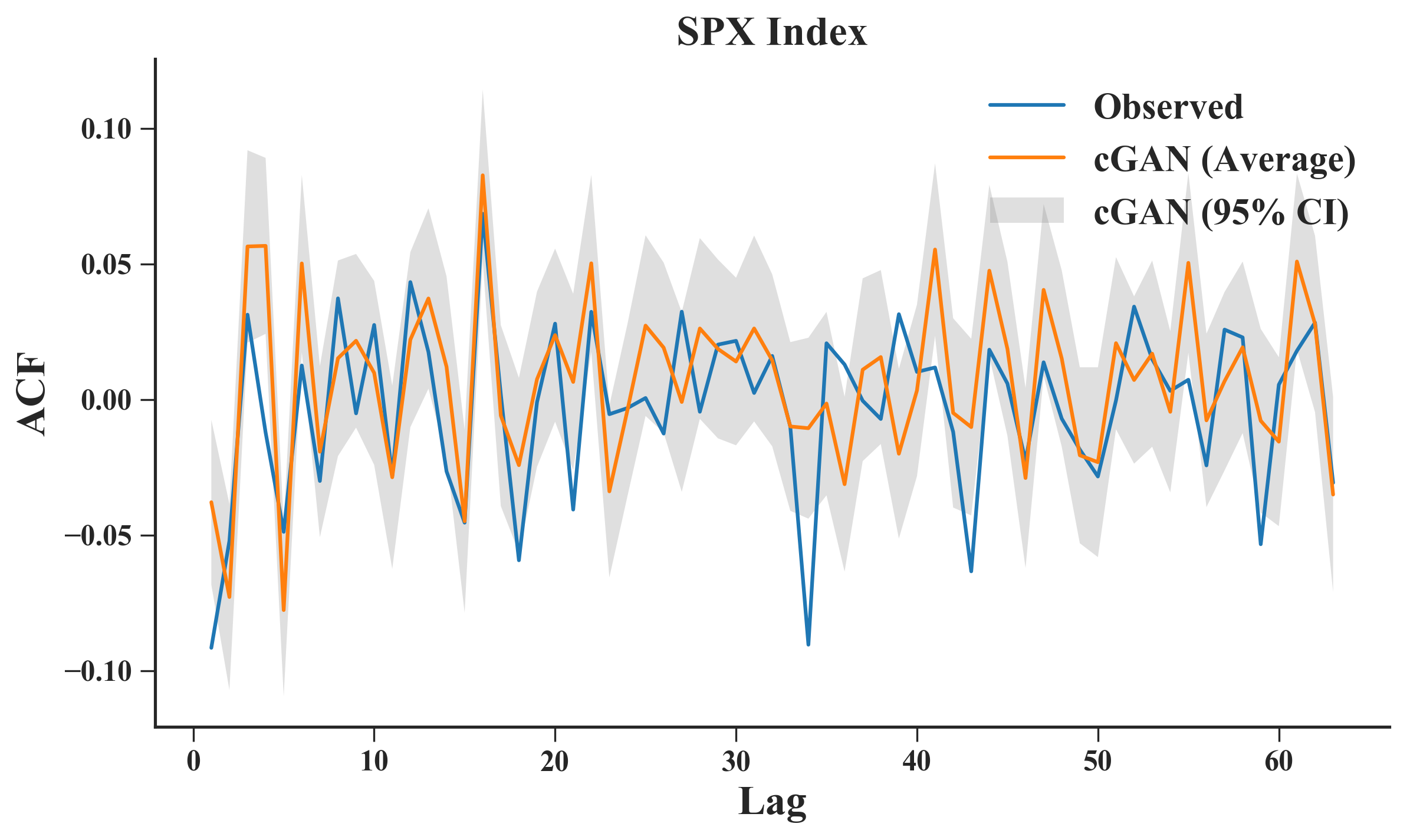}} \\ \subfloat[]{\includegraphics[width=\columnwidth]{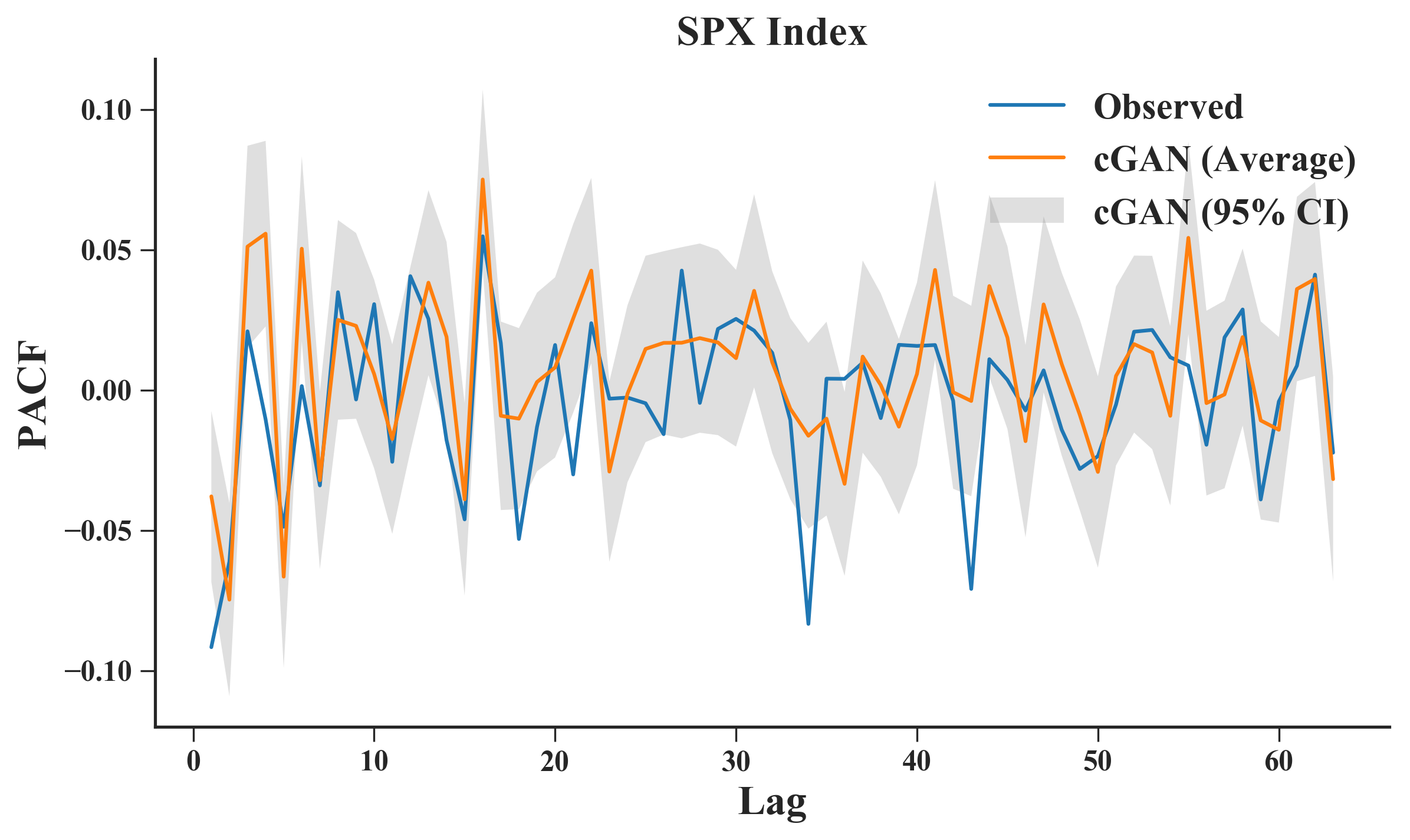}}
	\caption{Autocorrelations (a) and partial autocorrelations (b) for SPX Index using a cGAN with 1000 epochs.}
	\label{spx-cgan-corr}
\end{figure}

Clearly, with just a 200 epochs the samples generated do not resemble well the index, whilst with 1000 the results are closer. For SPX Index, not much improvement was observed after 1000-2000 epochs. Although the samples still appear similar to the index, some issues related to scaling and presence of overshoots in the prediction damaged the RMSE values. Figure \ref{spx-cgan-corr} look into the estimated autocorrelations (ACF) and partial autocorrelations (PACF) functions using samples of cGAN with 1000 epochs. With few exceptions, most of the observed ACF and PACF values were in the neighbourhood of the average of several cGAN samples, with the confidence interval (CI) covering most of the 63 lags; this anecdotal evidence suggest that our cGAN Training and Selection Algorithm is able to replicate to a certain extent some statistical properties of a time series, in particular its ACF and PACF. The proper evidence toward this last assertion are provided in our case studies.

A final note: we have adopted the Root Mean Square Error as the loss function between the generator samples and the actual data, however nothing limits the user to use another type of loss function. In the next two subsections we outline new applications that can be made using the cGAN generator: fine-tuning and combination of trading strategies.

\subsection{cGAN: Fine-Tuning of Trading Strategies}

Fine-tuning of trading strategies consists of identifying a suitable set of hyperparameters such that a desired goal is attained. This goal depends on what is the utility function $P$ that the quantitative analyst is targeting: outperformance during active trading, hedging a specific risk, reaching a certain level of risk-adjusted returns, and so on. This problem can be decomposed into two tasks -- model validation and hyperparameter optimization -- that are strongly connected. Using  \cite{bergstra2012random} as the initial step, given a(n):
\begin{itemize}
	\item finite set of examples $\mathbf{X}^{(train)}$ draw from a probability distribution $p_\mathbf{x} (\mathbf{x})$
	\item set of hyperparameters $\lambda \in \Lambda$, such as number of neurons, activation function of layer $j$, etc.
	\item utility function $P$ to measure a trading strategy $S_\lambda$ performance in face of new samples from $p_\mathbf{x} (\mathbf{x})$
	\item trading strategy $M_\lambda$ with parameters $\theta$ identifiable by an optimization of a training criterion, but only spotted after a certain $\lambda$ is fixed
\end{itemize}

mathematically, a trading strategy is fine-tuned properly if we are able to identify:
\begin{equation}
\lambda_\ast = arg\max_{\lambda \in \Lambda} \mathbb{E}_{\mathbf{x} \sim p_\mathbf{x}} [P(x; M_\lambda(\mathbf{X}^{(train)}))]
\end{equation}
that is, the optimal configuration for $ M_{\lambda_\ast}$ that maximizes the generalization of utility $P$. In reality, since drawing new examples from $p_\mathbf{x}$ is hard, and $\Lambda$ could be extremely large, most of the work in hyperparameter optimization and model validation is done by a double approximation:
\begin{eqnarray}
\lambda_\ast = arg\max_{\lambda \in \Lambda} \mathbb{E}_{\mathbf{x} \sim p_\mathbf{x}} [P(\mathbf{x}; M_\lambda(\mathbf{X}^{(train)}))] \approx \nonumber \\
arg\max_{\lambda \in \{\lambda_1, \lambda_2, ..., \lambda_m\}} \mathbb{E}_{\mathbf{x} \sim p_\mathbf{x}} [P(\mathbf{x}; M_\lambda(\mathbf{X}^{(train)}))] \label{approx1} \approx \\
arg\max_{\lambda \in \{\lambda_1, \lambda_2, ..., \lambda_n\}} mean_{\mathbf{x} \in \mathbf{X}^{(val)}} [P(\mathbf{x}; M_\lambda(\mathbf{X}^{(train)}))] \label{approx2}
\end{eqnarray}

\noindent the first approximation (eq. \ref{approx1}) is discretizing the search space of $\lambda$ (hopefully including $\lambda_\ast$) due to finite amount of computation. There are better ways to do this search, such as using Evolution Strategies \cite{loshchilov2016cma} or Bayesian Optimization \cite{eggensperger2013towards}, but this is not the focus of our work. The second approximation (eq. \ref{approx1}) replaces the expectation over sampling from $p_\mathbf{x}$, by an average over validation sets $\mathbf{X}^{(val)}$. Creating proper validation sets have been the focus of a substantial amount of research:
\begin{itemize}
	\item when $\mathbf{x}_1, ...,\mathbf{x}_n$ are sampled independently and identically distributed (iid), techniques such as k-fold-cross-validation \cite{bergmeir2018note} and iid bootstrap \cite{arlot2010survey} can be employed to create both $\mathbf{X}^{(train)}$ and $\mathbf{X}^{(val)}$.
	\item when $\mathbf{x}_1, ...,\mathbf{x}_n$ are not iid, then modifications have to be employed in order to create $\mathbf{X}^{(train)}$ and $\mathbf{X}^{(val)}$ adequately. In itself, it is an ongoing research topic, but we can mention the block-cross-validation and $hv$-block-cross-validation \cite{racine2000consistent}, sliding window \cite{arlot2010survey}, one-split (single holdout) \cite{arlot2010survey}, stationary bootstrap \cite{lahiri2013resampling}, as potential candidates.
\end{itemize}

\noindent in this work we follow a different thread: we attempt to build an approximation of drawing new examples from $p_\mathbf{x}$ using a cGAN. Algorithm \ref{calibration} outlines the steps followed to fine-tune a trading strategy using a cGAN generator. 

\begin{algorithm}
	\caption{Fine-tuning trading strategies using cGAN}\label{calibration}
	\begin{algorithmic}[1]
		\Procedure{cGAN}{$[y_1,...,y_T], params$} \\
		\Comment{train and select a cGAN for a time series $y_1,...,y_T$}
		\State \textbf{return} $G, D$
		\EndProcedure
		
		\Procedure{cGAN-Fine-tuning}{$G$, $[y_1,...,y_T]$, $B$}
		\For{$\lambda \gets \lambda_1, ..., \lambda_m$}
		\For{$b\gets 1, B$}
		\For{$t\gets p+1, T$}
		\State sample noise vector $\mathbf{z} \sim p_{\mathbf{z}} (\mathbf{z})$
		\State draw $y_t^\ast = G(\mathbf{z} | y_{t-1}...., y_{t-p})$
		\EndFor
		\State train data: $\mathbf{X}^{(train)} := (y_{p+1}^\ast, ..., y_{T-h}^\ast)$
		\State fit trading strategy: $ M_\lambda^{(b)} (\mathbf{X}^{(train)})$
		\State val data: $\mathbf{X}^{(val)} := (y_{T-h+1}^\ast, ..., y_{T}^\ast)$
		\State perf: $s_\lambda^{(b)} = P(\mathbf{X}^{(val)}; M_\lambda^{(b)} (\mathbf{X}^{(train)}))$
		\EndFor
		\State average: $perf(\lambda) = (1/B) \sum_{b=1}^B s_\lambda^{(b)}$
		\EndFor
		\State \textbf{return} $arg \max_{\lambda \in \{\lambda_1, \lambda_2, ..., \lambda_m\}} perf(\lambda)$
		\EndProcedure
	\end{algorithmic}
\end{algorithm}

Hence, we train a cGAN and use the generator $G$ to draw $B$ samples from the time series. For every sample, we perform an one-split to create $\mathbf{X}^{(train)}$ and $\mathbf{X}^{(val)}$, so that we are able to identify $M_\lambda$ parameters $\lambda$ and assess a set of hyperparameters $\lambda$. Following eq. (\ref{approx2}), we return the hyperparameter $\lambda_\ast$ that maximize the average performance across the cGAN samples. The one-split method has one parameter $h$ which sets the holdout set size; its value is specified in the experimental setting section. We compared our methodology results with other schemes that produce $\mathbf{X}^{(train)}$ and $\mathbf{X}^{(val)}$ from a given dataset. Next subsection outline another use of the cGAN generator: ensemble modelling.

\subsection{cGAN: Sampling and Aggregation}

Another potential use of cGAN is to build an ensemble of trading strategies, that is, using base learners that are individually "weak" (e.g. Classification and Regression Tree), but when aggregated can outcompete other "strong" learners (e.g., Support Vector Machines). Notorious instantiations of this principle are Random Forest, Gradient Boosting Trees, etc., techniques that make use of Bagging, Boosting or Stacking \cite{friedman2001elements,efron2016computer}. In our case, the closest parallel we can draw to cGAN Sampling and Aggregation is Bagging. Algorithm \ref{cganning} shows this method. After have trained and selected a cGAN, we repeatedly draw a cGAN sample and train a base learner; having proceed this way for $b=1,...,B$ steps we return the whole set of base models as an ensemble.

\begin{algorithm}
	\caption{cGAN Sampling and Aggregation}\label{cganning}
	\begin{algorithmic}[1]
		\Procedure{cGAN}{$[y_1,...,y_T], params$} \\
		\Comment{train and select a cGAN for a time series $y_1,...,y_T$}
		\State \textbf{return} $G, D$
		\EndProcedure
		
		\Procedure{cGAN-Sample-Agg}{$G$, $[y_1,...,y_T]$, $B$}
		\For{$b\gets 1, B$}
		\For{$t\gets p+1, T$}
		\State sample noise vector $\mathbf{z} \sim p_{\mathbf{z}} (\mathbf{z})$
		\State draw $y_t^\ast = G(\mathbf{z} | y_{t-1}...., y_{t-p})$
		\EndFor
		\State train base learner: $ M_\lambda^{(b)} (y_{p+1}^\ast, ..., y_T^\ast)$
		\EndFor 
		\State \textbf{return ensemble} $M_\lambda^{(1)}, ..., M_\lambda^{(B)}$
		\EndProcedure
	\end{algorithmic}
\end{algorithm}

An argument that is often used to show why this scheme work is the variance reduction lemma \cite{friedman2001elements}: let $\hat{Y}_1, ..., \hat{Y}_B$ be a set of base learners, each one trained using distinct samples draw repeatedly from the cGAN generator. Then, if we average their predictions and analyse its variance we have:
\begin{equation}
\mathbb{V}\Big[\frac{1}{B} \sum_{b=1}^{B} \hat{Y}_b\Big] = \frac{1}{B^2} \Big(\sum_{b=1}^B \mathbb{V}[\hat{Y}_b] + 2 \sum_{1\leq b \leq j \leq B}^B\mathbb{C}[\hat{Y}_b, \hat{Y}_j]\Big)
\end{equation}

\noindent if we assume, for analytical purpose, that $\mathbb{V}[\hat{Y}_b] = \sigma^2$ and $\mathbb{C}[Y_b, Y_j] = \rho \sigma^2$ for all $b$, that is, equal variance $\sigma^2$ and average correlation $\rho$, this expression simplifies to:
\begin{equation}
\mathbb{V}\Big[\frac{1}{B} \sum_{b=1}^{B} \hat{Y}_b\Big] = \sigma^2 \Big(\frac{1}{B} + \frac{B-1}{B} \rho \Big) \leq \sigma^2
\end{equation}

\noindent hence, we are able to reduce a base learner variance by averaging many slightly correlated predictors. By Bias-Variance trade-off \cite{friedman2001elements,efron2016computer}, the ensemble Mean Squared Error tend to be minimized, particularly when low bias and high variance base learners are used, such as deep Decision Trees. Diversification in the pool of predictors is the key factor; commonly it is implemented by taking $B$ iid bootstrap samples from a dataset. When dealing with time series, iid bootstrap can corrupt its autocorrelation structure, and taking $B$ stationary bootstrap samples \cite{lahiri2013resampling} is preferred. Bagging predictors using stationary bootstrap is, therefore, the appropriate benchmark to compare cGAN Sampling and Aggregation. The method that is able to produce $\hat{Y}_b$ and $\hat{Y}_j$ with low $\sigma^2$ and as slightly correlated as possible, will tend to outperform out-of-sample. A final note: one potential risk is that cGAN is unable to replicate well $p_{data}$. Therefore, thought the samples are more diverse they are also more "biased". This can make the base learners to miss patterns displayed in the real dataset, or even spot ones that did not existed in the first place.

\section{Experimental Setting}

\subsection{Datasets and Holdout Set}

\begin{figure}[h!]
	\centering
	\includegraphics[width=\linewidth]{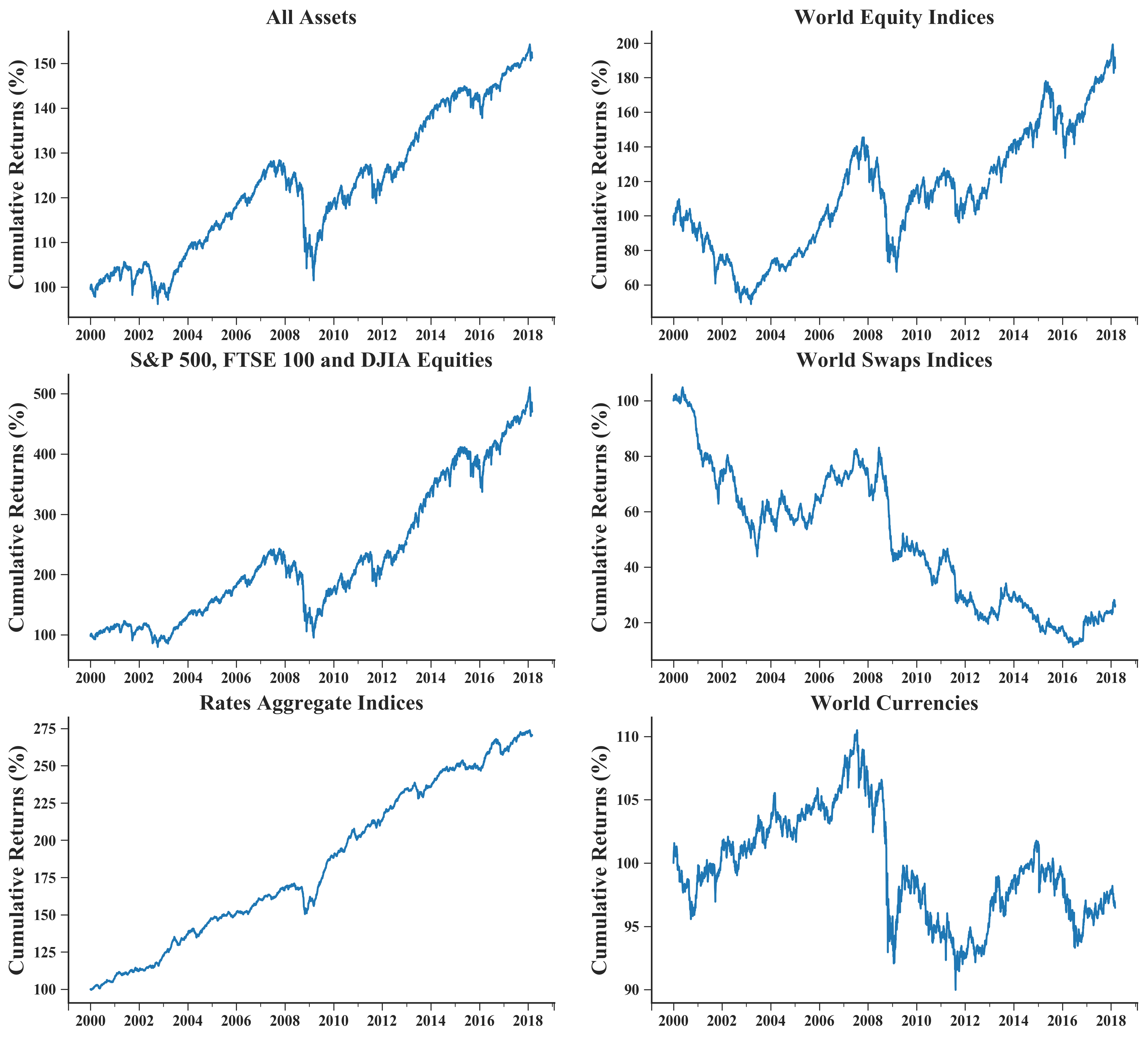} 
	\caption{Cumulative returns aggregated across asset pool. Before being averaged, each individual asset was volatility scaled to 10\%} \label{cum_returns_assets_class}
\end{figure}

Table \ref{datasets} presents aggregated statistics associated to the datasets used, whilst Figure \ref{cum_returns_assets_class} illustrates the cumulative returns per asset pool. We have considered three main asset classes during our evaluation: equities, currencies, and fixed income. The data was obtained from Bloomberg, with the full list of 579 assets tickers and period available at \texttt{\url{https://www.dropbox.com/s/08mjq7z49ybftqg/cgan_data_list.csv?dl=0}}. The typical time series started on 03/01/2000 and finished at 02/03/2018, with an average length of 4565 data points. We converted the raw prices in excess log returns, using a 3-month Libor rate as the benchmark rate.

\begin{table*}[h!]
	\centering
	\small
	\caption{Aggregated statistics of the assets used during our empirical evaluation.} \label{datasets}
	\begin{tabular}{p{3cm}|ccccccc}
		\hline
		\hline
		Asset Pool*	& N & Avg Return & Volatility & Sharpe ratio & Calmar ratio & Monthly skewness & VaR 95\% \\
		\hline
		All Assets & 579 & 0.0220 &	0.0453 &	0.4854 & 0.1051 &	-1.2169 &	-0.9443 \\
		World Equity Indices &	18 & 0.0152 & 0.0717 & 0.2114 & 0.0504 & -0.8170 &	-1.0048 \\
		S\&P 500, FTSE 100 and DJIA Equities & 491 &	0.0251 &	0.0525 &	0.4785 &	0.1127 &	-1.1133 &	-0.9517 \\
		World Swaps Indices & 48  & -0.0191 & 0.0446 & -0.4295 & -0.0458 & 0.0275 & -0.9753 \\
		Rates Aggregate Indices & 16 & 0.1220 & 0.0637 & 1.9135 & 0.5504 & -0.8227 &	-0.9440 \\
		World Currencies & 24 & -0.0025 & 0.0315 & -0.0798 & -0.0157 & -1.0052 & -0.8856 \\
		\hline
		\hline
	\end{tabular}
	\\ * Before being averaged, each individual asset was volatility scaled to 10\%
\end{table*}

\begin{figure}[h!]
	\centering
	\includegraphics[width=\linewidth]{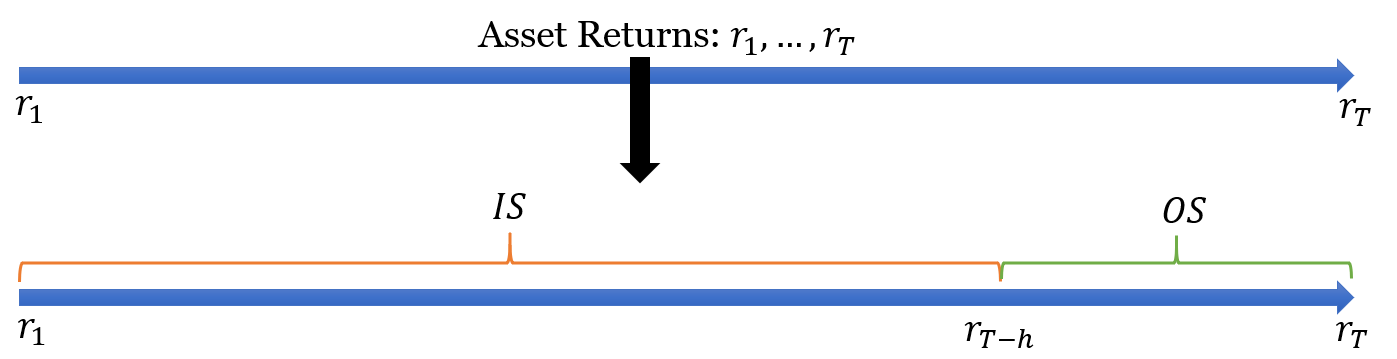}
	\caption{One-split/single holdout approach to assess all approaches in this work. During our experiments we have fixed $h=1260$ days $\approx 5$ years.} \label{onesplit}
\end{figure}

We have established a testing procedure to assess all the different approaches spanned in this research. Figure \ref{onesplit} summarize the whole procedure. The process start by splitting a sequence of returns $r_1, ..., r_T$ in a single in-sample/training ($IS$) and out-sample/testing/holdout ($OS$) set, with both sets sizes being determined by the trading horizon $h$. During our experiments we have fixed $h=1260$ days $\approx 5$ years. Every method used or cGAN trained tap only into the $IS$ data. Some methods, such as the other Model Validation schemes will create training and validation sequences, but all of them only based on $IS$ set. However, the data used to measure their success is the same: by computing a set of metrics using the fixed $OS$ set.

\subsection{Performance Metrics}

This subsection outline the utility function employed. We opted for a financial performance metric, instead of a generic metric based in prediction error. Low prediction error is a necessary, but not a sufficient condition to construct alpha-generating trading strategies \cite{acar2002advanced}. In this sense, we mainly reported Sharpe and Calmar ratios: metrics that combine risk and reward measures, and make different investment strategies comparable \cite{young1991calmar,sharpe1994sharpe,eling2007does}. These metrics can be defined as:

\begin{equation}
SR = \frac{\bar{R}^{(M)}}{\sigma_R^{(M)}} \ \ \mathrm{and} \ \ CR = \frac{\bar{R}^{(M)}}{-MDD(R^{(M)})}
\end{equation}

\noindent where $\bar{R}^{(M)}$ is the strategy average annualized excess returns, $\sigma_R^{(M)}$ is it volatility, and $MDD(R^{(M)})$ is the strategy maximum drawdown. All of them are calculated using the strategy instantaneous returns $r^{(M)}_t = r_t \cdot f(\hat{r}_t)$ as the building block. In this case, $f(\hat{r}_t)$ is our trading signal, a transformation $f$ of the estimated returns $\hat{r}_t$ outputted by a predictive model. We opted to use an identity function for $f(\hat{r}_t) = \hat{r}_t$ so we can avoid having another layer of hyperparameters; in practice an user can select another transformation.

Finally, it should be mentioned that most of our reported results are aggregated across all 579 assets. Although the ratios provide a common basis to compare strategies with different risk profiles, still, the asset pool is quite diverse and is affected by outliers. Hence, we opted for robust statistics (median, mean absolute deviation, quantiles, ranks, etc.) to compare and statistically test the different hypothesis\footnote{The readers interested to understand more about the nonparametric statistical tests used in this work -- Friedman, Holm Correction and Wilcoxon rank-sum test -- should consult this reference \cite{derrac2011practical}.}.

\subsection{cGAN Configuration}

Table \ref{cgan-configs} outlines the three different architectures used for $G$ and $D$ of a cGAN. Since the main variation is the number of neurons used, we abbreviated their names across the cases as cGAN-Small, cGAN-Medium and cGAN-Large. This variation will allow us to check how different architecture sizes perform across the benchmarks.

\begin{table}[h!]
	\centering
	\footnotesize
	\caption{Configurations used to train and select the cGANs.} \label{cgan-configs}
	\begin{tabular}{p{4cm}p{4cm}}
		\hline
		\hline
		\multicolumn{1}{c}{\textbf{Abbreviation}} & \multicolumn{1}{c}{\textbf{Individual Configuration}} \\
		\hline
		cGAN-Small ($G$, $D$) & number of neurons = (5, 5) \\
		cGAN-Medium ($G$, $D$) & number of neurons = (100, 100) \\
		cGAN-Large ($G$, $D$) & number of neurons = (500, 500)\\
		\hline
		\hline
		\hline
		\multicolumn{1}{c}{\textbf{Other Configuration}} & \multicolumn{1}{c}{\textbf{Values}} \\ \hline
		Architecture & Multilayer Perceptron \\
		Number of hidden layers & 1 \\
		Hidden layer activation function & rectified linear \\
		$G$ Output layer activation function & linear \\
		$D$ Output layer activation function & sigmoid \\
		Epochs & 20000 \\
		Batch Size & 252 \\
		Solver & Stochastic Gradient Descent \\
		Solver Parameters & learning rate = 0.01 \\
		Conditional info &  $r_{t-1}, r_{t-2}, ..., r_{t-252}$ ($p=252$) \\
		Noise prior $p_{\mathbf{z}}(\mathbf{z})$ & $N(0, 1)$ \\
		Noise dimension $dim(\mathbf{z})$ & 252 \\
		Snapshot frequency ($snap$) & 200 \\
		Number of samples for evaluation & $C =$ 50 \\
		Input features scaling function & $Z$-score (standardization) \\
		Target scaling function & $Z$-score (standardization) \\
		\hline
		\hline
	\end{tabular}
\end{table}

After a few initial runs, we opted for Stochastic Gradient Descent with learning rate of 0.01 and batch size of 252 as the optimization algorithm. Input features and target were scaled using a z-score function to ease the training process. We selected the right cGAN to use by taking snapshots every 200 iterations ($snap=200$), drawing and evaluating 50 samples per generator along 20000 epochs.  Finally, we used 252 consecutive lags as conditional information (around one year) with the noise prior (Standard Normal - $N(0,1)$) wielding the same dimension of the conditional input; we did it to increase the chance to create a more diverse pool of examples, as well as to make it harder for the generator to omit/nullify the weights associated with this part of the input layer.

\subsection{Case I: Combination of Trading Strategies}

\subsubsection{Overview}

This case evaluates the success of different combination of trading strategies. In this sense, Algorithm \ref{combination-algo} presents the main loop used for cGANs and Stationary Bootstrap. First step is to resample the actual returns $RS(r_1, ...,r_{T-h})$ using Stationary Bootstrap or cGAN, creating a new sequence of returns $\{r_1^\ast, ...,r_{T-h}^\ast\} = \mathbf{X}^{(train)}$ set. We then proceed as usual: use $\mathbf{X}^{(train)}$ to train a base learner $M_\lambda^{(b)}$, and add it to the ensemble set $ES$ All of these steps are repeated $B$ times. Finally, we can propagate the $OS$ feature set through the ensemble $ES$, get the aggregated prediction, and compute its performance within this holdout set.

\begin{algorithm}
	\caption{Generic loop for combination of strategies}\label{combination-algo}
	\begin{algorithmic}[1]	
		\For{$b\gets 1, B$}
		\State $\mathbf{X}^{(train)} := \{r_1^\ast, ...,r_{T-h}^\ast\} = RS(r_1, ...,r_{T-h})$
		\State fit trading strategy: $ M_\lambda^{(b)} (\mathbf{X}^{(train)})$
		\State add to ensemble: $ES \gets M_\lambda^{(b)} (\mathbf{X}^{(train)})$
		\EndFor
		\State test ensemble: $P(r_{T-h}, ...,r_{T}; Agg(ES))$
	\end{algorithmic}
\end{algorithm}

\subsubsection{Methods and Parameters}

Table \ref{config-combination} presents the instantiations of $RS$, $M_\lambda$, and $Agg$ of Algorithm \ref{combination-algo}. The main competing method is the Stationary Bootstrap; for all $RS$ schemes, we have taken different number of resamples $B$, so that we could compare the efficiency for different sizes of the ensemble. We used two main base learners: a deep Regression Tree and a large Multilayer Perceptron. The main idea was to follow the usual principle of using low bias and high variance base learners. We employed a fixed feature set of 252 consecutive lags, and averaged the prediction of all members. Therefore, we can describe the main hypothesis as: {\em which resampling scheme $RS$ is able to create a set of trading strategies $ES =\{M_\lambda^{(1)}, ..., M_\lambda^{(B)}\}$ that in aggregate manage to outcompete during the $OS$ period?}

\begin{table}[h!]
	\centering
	\footnotesize
	\caption{Main configuration used for Case I: Combination of Trading Strategies.} \label{config-combination}
	\begin{tabular}{p{4cm}|p{4cm}}
		\hline
		\hline
		\multicolumn{1}{c}{\textbf{Resampling Scheme ($RS$)}} & 		\multicolumn{1}{c}{\textbf{Parameters}} \\
		\hline
		Stationary Bootstrap \cite{lahiri2013resampling} & $B = \{20, 100, 500\}$ samples and \ \ \ \ \ \ \  \ \ \ \ \ \ \ \ \ \ \ \ \ block size = 20 \\
		cGAN-Small & $B = \{20, 100, 500\}$ samples \\
		cGAN-Medium & $B = \{20, 100, 500\}$ samples \\
		cGAN-Large & $B = \{20, 100, 500\}$ samples \\
		\hline
		\hline
		\multicolumn{1}{c}{\textbf{Trading Strategy ($M_\lambda$)}} & \multicolumn{1}{c}{\textbf{Hyperparameters ($\lambda$)}} \\ 
		\hline
		Regression Tree (Reg Tree) \cite{efron2016computer} & unlimited depth, with minimum number of samples required to split an internal node of 2 \\
		\hline
		Multilayer Perceptron (MLP) \cite{efron2016computer} & number of neurons = $\{200\}$, weight decay = $\{0.00001\}$ and \ \ \ \ \ activation function = $\{tanh\}$ \\
		\hline
		\hline
		\multicolumn{1}{c}{\textbf{Other details}} & 		\multicolumn{1}{c}{\textbf{Values}} \\
		\hline
		Number of lags used as features & $r_{t-1}, r_{t-2}, ..., r_{t-252}$ \\
		Aggregation function ($Agg$) & Mean \\
		\hline
		\hline
	\end{tabular}
\end{table}

\subsection{Case II: Fine-tuning of Trading Strategies}

\subsubsection{Overview}

This case evaluates the success of different fine-tuning strategies, in particular those that create $\mathbf{X}^{(train)}$ and $\mathbf{X}^{(val)}$ sets for time series. In this sense, Algorithm \ref{finetuning-algo} presents an unified loop used regardless of the methodology employed: from data splitting, hyperparameter selection, and performance calculation.

\begin{algorithm}
	\caption{Generic loop for fine-tuning of trading strategies}\label{finetuning-algo}
	\begin{algorithmic}[1]	
		\For{$b\gets 1, B$} \Comment{All training and validation folds}
		\State $\mathbf{X}^{(train)}, \mathbf{X}^{(val)} := MV(r_1, ...,r_{T-h})$
		\For{$\lambda \gets \lambda_1, ..., \lambda_m$}
		\State fit trading strategy: $ M_\lambda^{(b)} (\mathbf{X}^{(train)})$
		\State check strategy: $s_\lambda^{(b)} = P(\mathbf{X}^{(val)}; M_\lambda^{(b)} (\mathbf{X}^{(train)}))$
		\EndFor
		\EndFor
		\For{$\lambda \gets \lambda_1, ..., \lambda_m$}
		\State average across sets: $perf(\lambda) = (1/B) \sum_{b=1}^B s_\lambda^{(b)}$
		\EndFor
		\State \textbf{opt hyperparam}: $\lambda^\ast := arg \max_{\lambda \in \{\lambda_1, \lambda_2, ..., \lambda_m\}} perf(\lambda)$
		\State \textbf{fit trading strategy}: $M_{\lambda^\ast} (\mathbf{X}^{(train)}:= r_1, ...,r_{T-h})$
		\State \textbf{test trading strategy}: $P(r_{T-h}, ...,r_{T}; M_\lambda^{(b)} (\mathbf{X}^{(train)}))$
	\end{algorithmic}
\end{algorithm}

It start by splitting the $IS = \{r_1, ...,r_{T-h}\}$ set in  $\mathbf{X}^{(train)}$ and $\mathbf{X}^{(val)}$ using a Model Validation methodology $MV$ -- one-split, stationary bootstrap, cGAN, etc. Then, for every hyperparameter $\lambda_1, ..., \lambda_m$, we fit a trading strategy (e.g., Multi-layer Perceptron - $M_\lambda^{(b)}$) that aims to predict $r_t$ using lagged information $r_{t-1}, ..., r_{t-p}$ as the feature set. We check the strategy performance $s_\lambda^{(b)}$ using a validation set $\mathbf{X}^{(val)}$ and an utility function $P$ (e.g., Sharpe ratio). This process is repeated for all training and validation sets ($B$). Then, we measure the worthiness of a hyperparameter $\lambda$ (e.g., (number of neurons, weight decay) = (20, 0.05)) by averaging its performance across the validation folds $perf(\lambda)$; the optimal configuration is the one that maximizes the expected utility. Using this hyperparameter, a final model is fitted and tested using $OS$ set. 

\subsubsection{Methods and Parameters}

Table \ref{config-finetuning} presents the instantiations of $MV$, $M_\lambda$, $\lambda$ and $P$ of Algorithm \ref{finetuning-algo}. 

\begin{table}[h!]
	\centering
	\footnotesize
	\caption{Main configuration used for Case II: Fine-tuning of trading strategies.} \label{config-finetuning}
	\begin{tabular}{p{4cm}|p{4cm}}
		\hline
		\hline
		\multicolumn{1}{c}{\textbf{Model Validation ($MV$)}} & 	\multicolumn{1}{c}{\textbf{ Parameters}} \\
		\hline
		Naive ($\mathbf{X}^{(val)} = \mathbf{X}^{(train)}$) & - \\
		Sliding window \cite{arlot2010survey} & stride and window sizes $= 252$ days \\
		Block cross-validation \cite{racine2000consistent} & block size $= 252$ days \\
		hv-Block cross-validation \cite{racine2000consistent} & block size = 252 days and \ \ \ \ \ \ \ \ \ \ \ \ \ \ \ gap size = 10 days \\
		One-split/Holdout/Single split \cite{arlot2010survey} & $\mathbf{X}^{(val)}$ = last 1260 days \\
		k-fold cross-validation \cite{bergmeir2018note} & $k$ = 10 folds \\
		Stationary Bootstrap \cite{lahiri2013resampling} & $B$ = 100 samples and \ \ \ \ \ \ \  \ \ \ \ \ \ \ \ \ \ \ \ \ block size = 20 \\
		cGAN-Small & $B=$ 100 samples \\
		cGAN-Medium & $B=$ 100 samples \\
		cGAN-Large & $B=$ 100 samples \\
		\hline
		\hline
		\multicolumn{1}{c}{\textbf{Trading Strategy ($M_\lambda$)}} & \multicolumn{1}{c}{\textbf{Hyperparameters ($\lambda$)}} \\ 
		\hline
		Gradient Boosting Trees (GBT) \cite{efron2016computer} & number of trees = $\{50, 100, 200\}$, learning rate = $\{0.0001, 0.001$, $0.01, 0.1, 1.0\}$ and maximum depth = $\{1, 3, 5\}$ \\
		\hline
		Multilayer Perceptron (MLP) \cite{efron2016computer} & neurons = $\{20, 50, 100, 200\}$, weight decay = $\{0.001, 0.01$, $0.1, 1.0\}$ and activation function = $\{tanh\}$ \\
		\hline
		Ridge Regression (Ridge) \cite{efron2016computer} & shrinkage = $\{0.00001, 0.00005$, $0.0001, 0.0005$, $0.001, 0.005$, $0.01, 0.05$, $0.1, 0.5, 1.0\}$ \\
		\hline
		\hline
		\multicolumn{1}{c}{\textbf{Other details}} & 		\multicolumn{1}{c}{\textbf{Values}} \\
		\hline
		Number of lags used as features & $r_{t-1}, r_{t-2}, ..., r_{t-252}$ \\
		Hyperparameter search & Grid-search or Exhaustive search \\	
		Utility function $P$ & Sharpe ratio \\
		\hline
		\hline
	\end{tabular}
\end{table}

Apart from the three different architectures of cGANs, the competing methods to cGAN for fine-tuning trading strategies are: naive (training and validation sets are equal), one-split and sliding window; block, hv-block and k-fold cross-validation; stationary bootstrap. Hence, the main hypothesis is: {\em given a trading strategy $M_\lambda$, which $MV$ mechanism is able to uncover the best configuration $\lambda$ to apply during the $OS$ period?} We search for an answer to this hypothesis using linear and nonlinear trading strategies (Ridge Regression, Gradient Boosting Trees and Multilayer Perceptron). We used the Sharpe ratio as the utility function, grid-search as the hyperparameter search method, and a fixed feature set consisting of 252 consecutive lags.

\section{Case Studies}

\subsection{Case I: Combination of Trading Strategies}

Table \ref{main-results-case2-median} presents the median and mean absolute deviation (MAD - in brackets) results of ensemble strategies in the $OS$ set. Starting with Regression Tree (Reg Tree), we observe that the median Sharpe and Calmar ratios of cGAN-Large was higher across distinct number of base learners ($B = 20, 100, 500$). In fact, it was already twice as much of Stationary Bootstrap (Stat Boot), even when the number of samples was smaller ($B=20$); after this point some gain can still be obtained, but it seems that most of the diversification effect had already been realised. A different picture can be draw for Multilayer Perceptron (MLP): in this case Stat Boot produced better median Sharpe and Calmar ratios across the assets, with some exceptions when $B=20$. 

Looking into cGAN results, often the configuration cGAN-Large performed better, whilst in the other side of the spectrum cGAN-Small underperforming. Overall, our results suggest that using a high capacity MLP as the Generator/Discriminator helps to produce a Resampling Strategy that favours the training of base learners. We also reported the Root Mean Square Error (RMSE) since it is usual to report it for Ensemble Strategy. Numerically, they were very similar, nevertheless cGAN-Medium obtained the best values across $B$ and trading strategies.

\begin{table*}[h!]
	\centering
	\scriptsize
	\caption{Median and Mean Absolute Deviation (MAD) results of Trading and Ensemble Strategies on the $OS$ set.} \label{main-results-case2-median}
	\begin{tabular}{c|c|c|cccc}
		\hline
		\hline
		\multirow{2}{*}{Trad Strat} & \multirow{2}{*}{Metric} & \multirow{2}{*}{$B$} & \multicolumn{4}{c}{Ensemble Strategy} \\
		& & & Stat Boot & cGAN-Small & cGAN-Medium & cGAN-Large \\
		\hline
		\multirow{9}{*}{Reg Tree} & \multirow{3}{*}{Sharpe} & 20 & 0.042560 (0.380039) & 0.053867 (0.378896) & 0.044741 (0.380228) & \textbf{0.080540} (0.360695) \\
		&  & 100 & 0.062837 (0.378920) & 0.058820 (0.387749) & 0.030588 (0.390575) & \textbf{0.086423} (0.406171) \\
		&  & 500 & 0.074116 (0.397212) & 0.067905 (0.392788) & 0.072071 (0.392382) & \textbf{0.098094} (0.424621) \\ \cline{2-7}
		
		& \multirow{3}{*}{Calmar} & 20 & 0.019442 (0.230641) & 0.022619 (0.201044) & 0.018987 (0.200625) & \textbf{0.035473} (0.191353) \\
		&  & 100 & 0.027235 (0.241023) & 0.024254 (0.209783) & 0.011890 (0.201523) & \textbf{0.036523} (0.239046) \\
		&  & 500 & 0.034422 (0.266419) & 0.031174 (0.212710) & 0.032761 (0.221514) & \textbf{0.042232} (0.251194) \\ \cline{2-7}
		
		& \multirow{3}{*}{RMSE} & 20 & 0.014397 (0.005570) & 0.014561 (0.005604) & \textbf{0.014289} (0.005414) & 0.014411 (0.005432) \\
		&  & 100 & 0.014096 (0.005486) & 0.014281 (0.005545) & \textbf{0.013988} (0.005357) & 0.014099 (0.005373) \\
		&  & 500 & 0.014035 (0.005470) & 0.014203 (0.005531) & \textbf{0.013912} (0.005346) & 0.014033 (0.005361) \\ \cline{2-7}
		\hline
		\multirow{9}{*}{MLP} & \multirow{3}{*}{Sharpe} & 20 & 0.080722 (0.390515) & 0.079428 (0.416847) & \textbf{0.087960} (0.393913) & 0.069866 (0.398197) \\
		&  & 100 & \textbf{0.097576} (0.382028) & 0.063012 (0.415537) & 0.091344 (0.397506) & 0.087216 (0.414697) \\
		&  & 500 & \textbf{0.092262} (0.390161) & 0.059344 (0.415700) & 0.073652 (0.389588) & 0.085333 (0.414096) \\ \cline{2-7}
		
		& \multirow{3}{*}{Calmar} & 20 & 0.035525 (0.223141) & 0.030805 (0.217727) & \textbf{0.037877} (0.219139) & 0.031145 (0.214533) \\
		&  & 100 & \textbf{0.045916} (0.227827) & 0.023479 (0.223602) & 0.040718 (0.217648) & 0.040572 (0.223359) \\
		&  & 500 & \textbf{0.038678} (0.237459) & 0.024014 (0.225413) & 0.035688 (0.215691) & 0.035885 (0.222552) \\ \cline{2-7}
		
		& \multirow{3}{*}{RMSE} & 20 & 0.014030 (0.005416) & 0.013999 (0.005408) & \textbf{0.013910} (0.005345) & 0.014055 (0.005369) \\
		&  & 100 & 0.013924 (0.005399) & 0.013973 (0.005403) & \textbf{0.013878} (0.005339) & 0.014028 (0.005363) \\
		&  & 500 & 0.013924 (0.005400) & 0.013974 (0.005402) & \textbf{0.013887} (0.005337) & 0.014033 (0.005362) \\
		\hline
		\hline
		
	\end{tabular}
\end{table*}

\begin{figure}[h!]
	\centering
	\subfloat[]{\includegraphics[width=\columnwidth]{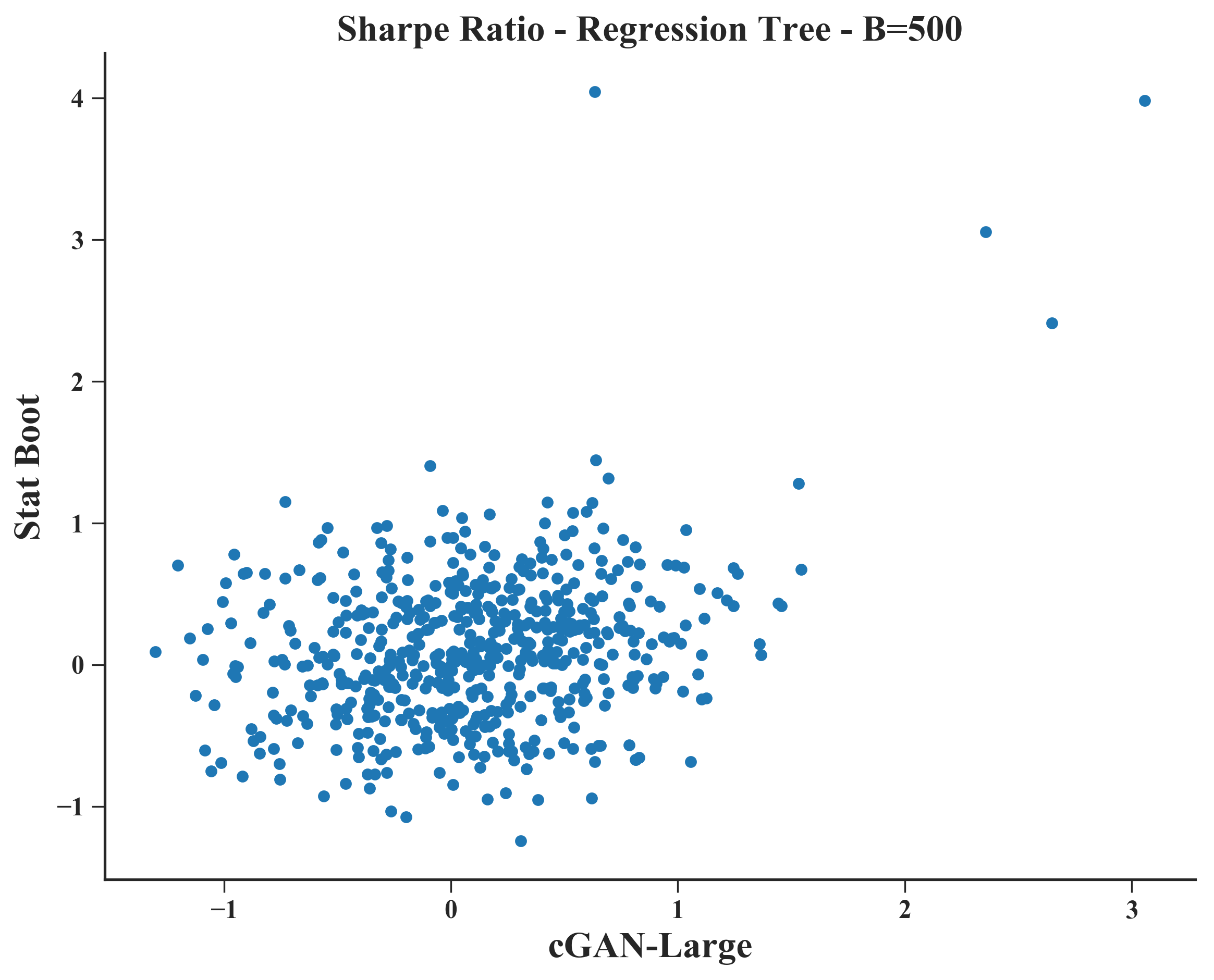}} \\
	\subfloat[]{\includegraphics[width=\columnwidth]{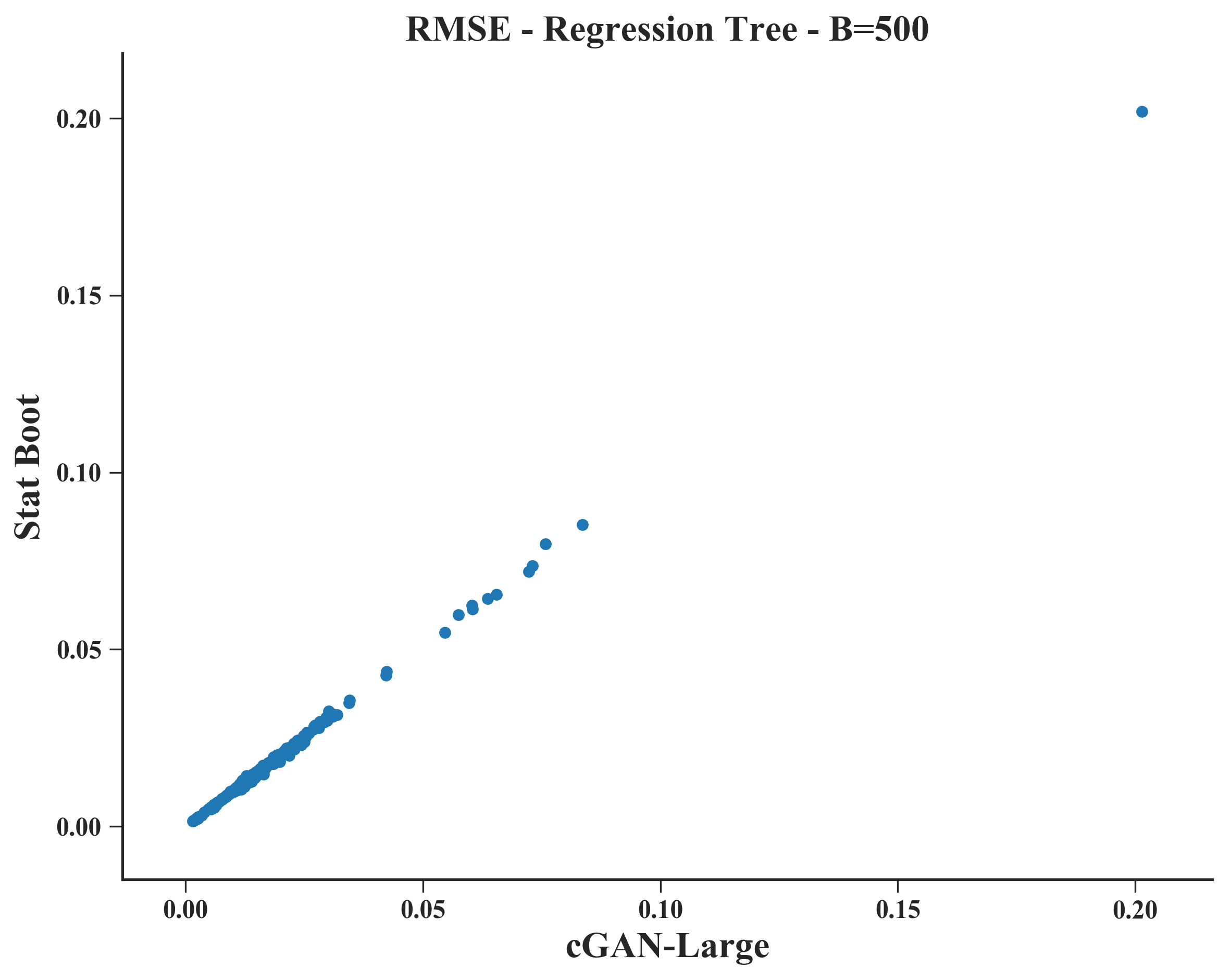}}
	\caption{Scatterplot of Sharpe ratio (a) and RMSE (b) values obtained using cGAN-Large and Stat Boot across 579 assets.} \label{cgan-statboot-metrics}
\end{figure}

Except for RMSE, MAD values were high for Sharpe and Calmar ratios across the different combinations. Hence in aggregate, any numerical difference can become imperceptible from statistical lens. Table \ref{main-results-case2-pvalues} shows if some of the differences raised about the values in Table \ref{main-results-case2-median}, only between cGAN-Large and Stat Boot, are statistically significant or not. Overall, apart from RMSE, p-values of Wilcoxon rank-sum test were in general above 0.05 (significance level adopted), meaning that the differences observed were not substantial across models, number of samples, and Sharpe or Calmar ratios.

\begin{table}[h!]
	\centering
	\footnotesize
	\caption{p-values of Wilcoxon rank-sum test comparing cGAN-Large with Stationary Bootstrap results.} \label{main-results-case2-pvalues}
	\begin{tabular}{c|c|ccc}
		\hline
		\hline
		\multirow{2}{*}{Trad Strat} & \multirow{2}{*}{Metric} & \multicolumn{3}{c}{$B$} \\ 
		& & 20 & 100 & 500 \\
		\hline
		\multirow{3}{*}{Reg Tree} & Sharpe & 0.2440 & 0.3432 & 0.7832 \\
		
		& Calmar & 0.8495 & 0.2958 & 0.9106 \\
		
		& RMSE & \textbf{0.0456} & \textbf{$<$ 0.0001} & \textbf{$<$ 0.0001} \\
		
		\hline
		\multirow{3}{*}{MLP} & Sharpe & 0.7941 & 0.3994 & 0.4295 \\
		
		& Calmar & 0.8119 & 0.4805 & 0.4053 \\
		
		& RMSE & 0.7973 & \textbf{0.0043} & \textbf{0.0007} \\
		\hline
		\hline
		
	\end{tabular}
\end{table}

In principle, so far it seems that there is little difference between cGAN-Large and Stat Boot, across models, metrics and number of samples. However, this equivalence in aggregate often do not manifest itself at the micro level. Figure \ref{cgan-statboot-metrics} presents this analysis: plotting the Sharpe ratio and RMSE obtained for every asset using cGAN-Large and Stat Boot ($B=500$). For RMSE there is an almost perfect correlation -- when cGAN-Large thrives, Stationary Bootstrap also do, with the converse also holding. However, a different phenomena occurs for Sharpe ratio: apart from a few outliers that skewed the correlation (0.407733), when Stat Boot fails to deliver reasonable results, cGAN-Large can provide a feasible alternative for combining weak signals. This complementarity, not perceived when looked in aggregate, can be an asset for the quantitative analyst in its pursuit to build alpha generating strategies.

\begin{figure*}[h!]
	\centering
	\textbf{Regression Tree \ \ \ \ \ \ \ \ \ \ \ \ \ \ \ \ \ \ \ \  \ \ \ \ \ \ \ \ \ \ \ \ \ \ \ \ \ \ \ \ \ \ \ \ \ \ \ Multilayer Perceptron}\par
	\subfloat[]{\includegraphics[width=.85\columnwidth]{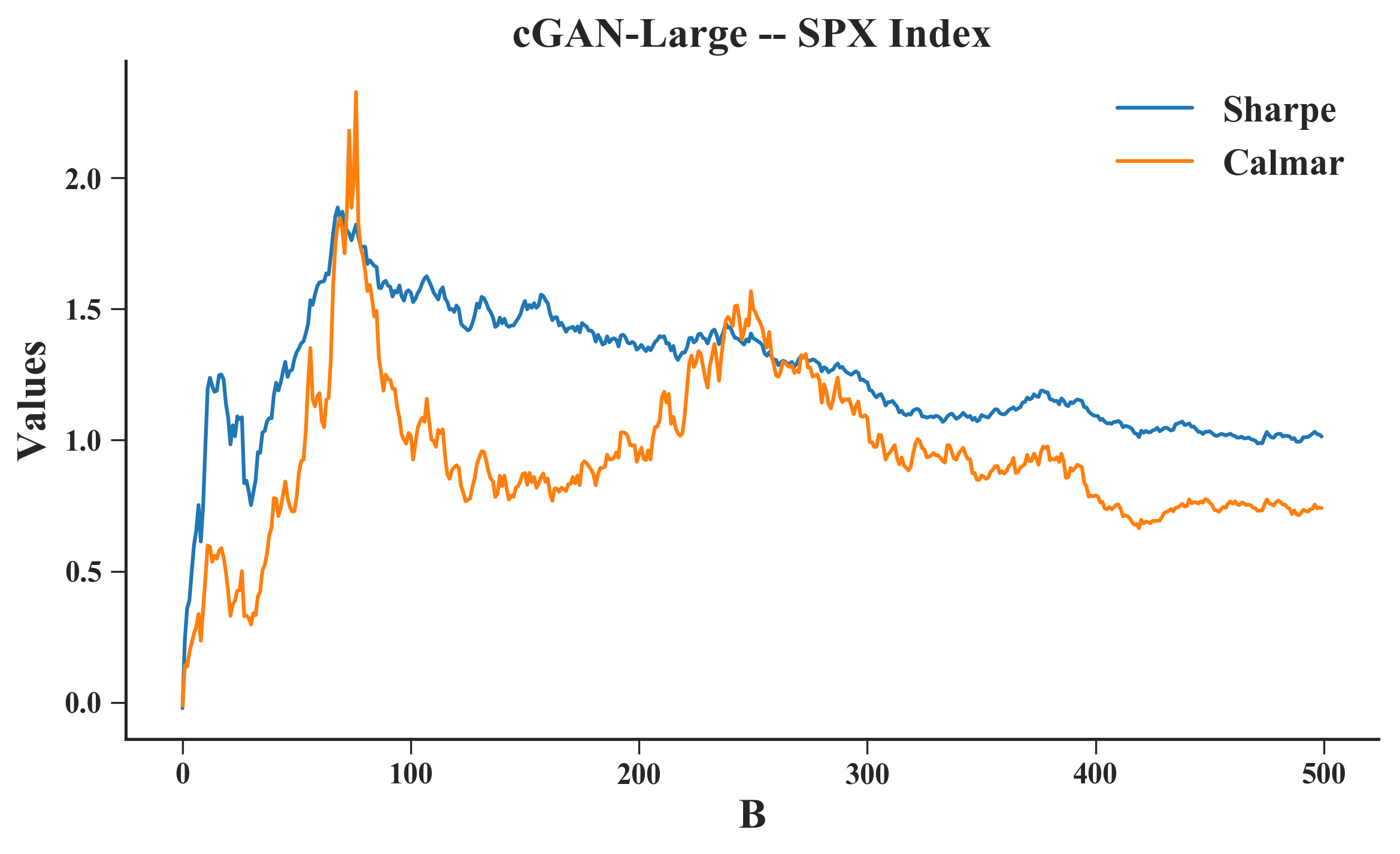}} \subfloat[]{\includegraphics[width=.85\columnwidth]{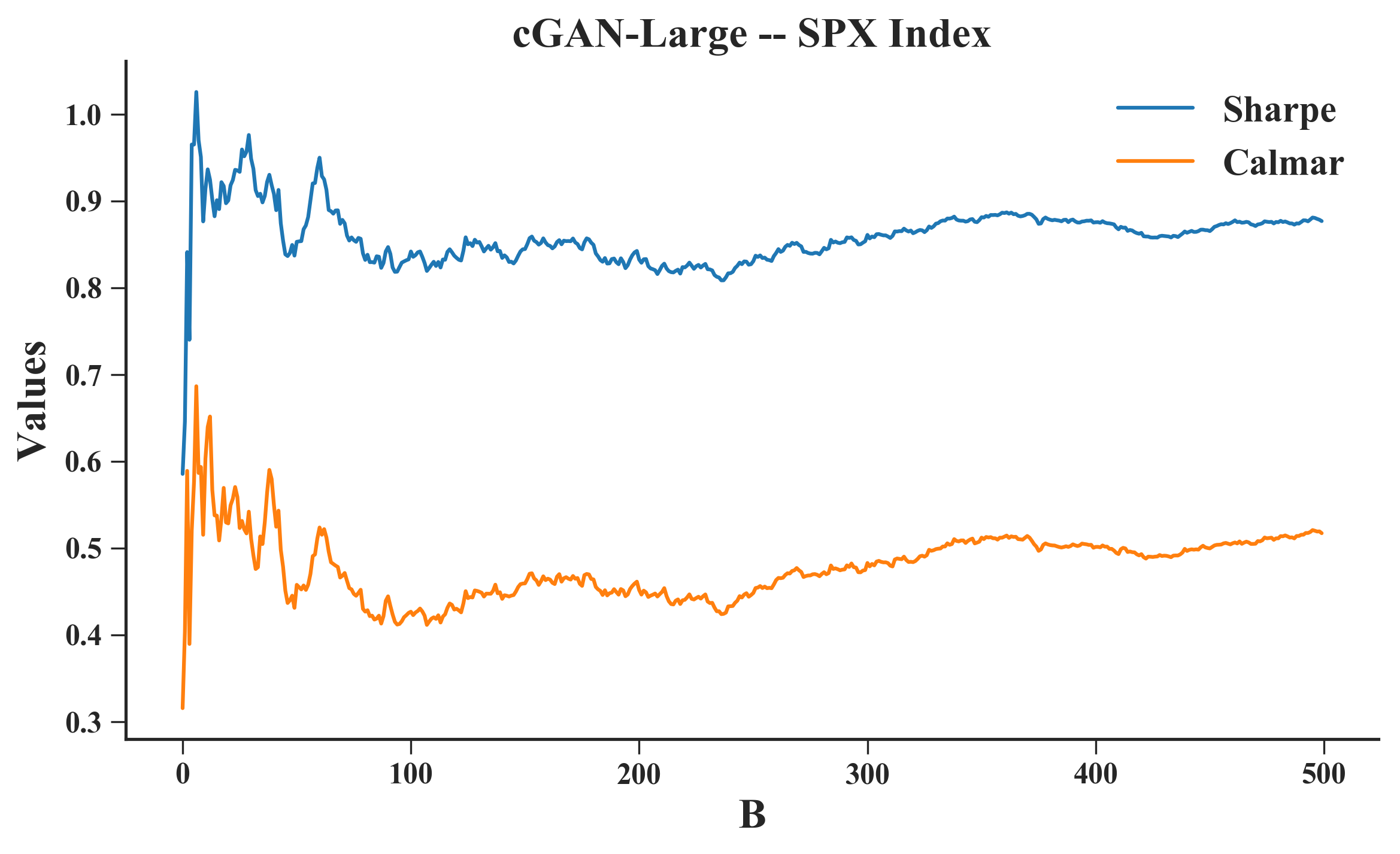}} \\
	\subfloat[]{\includegraphics[width=.85\columnwidth]{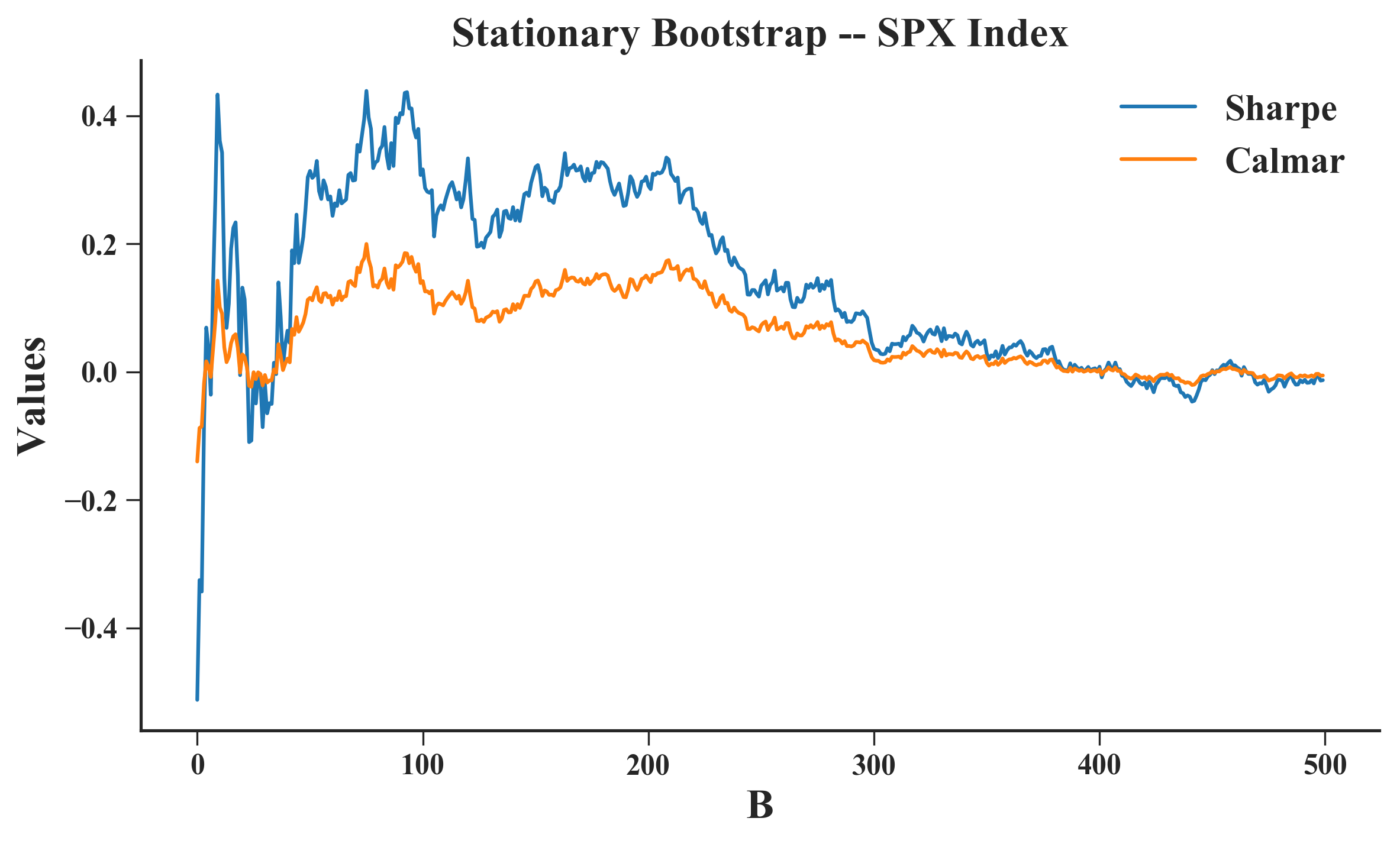}} \subfloat[]{\includegraphics[width=.85\columnwidth]{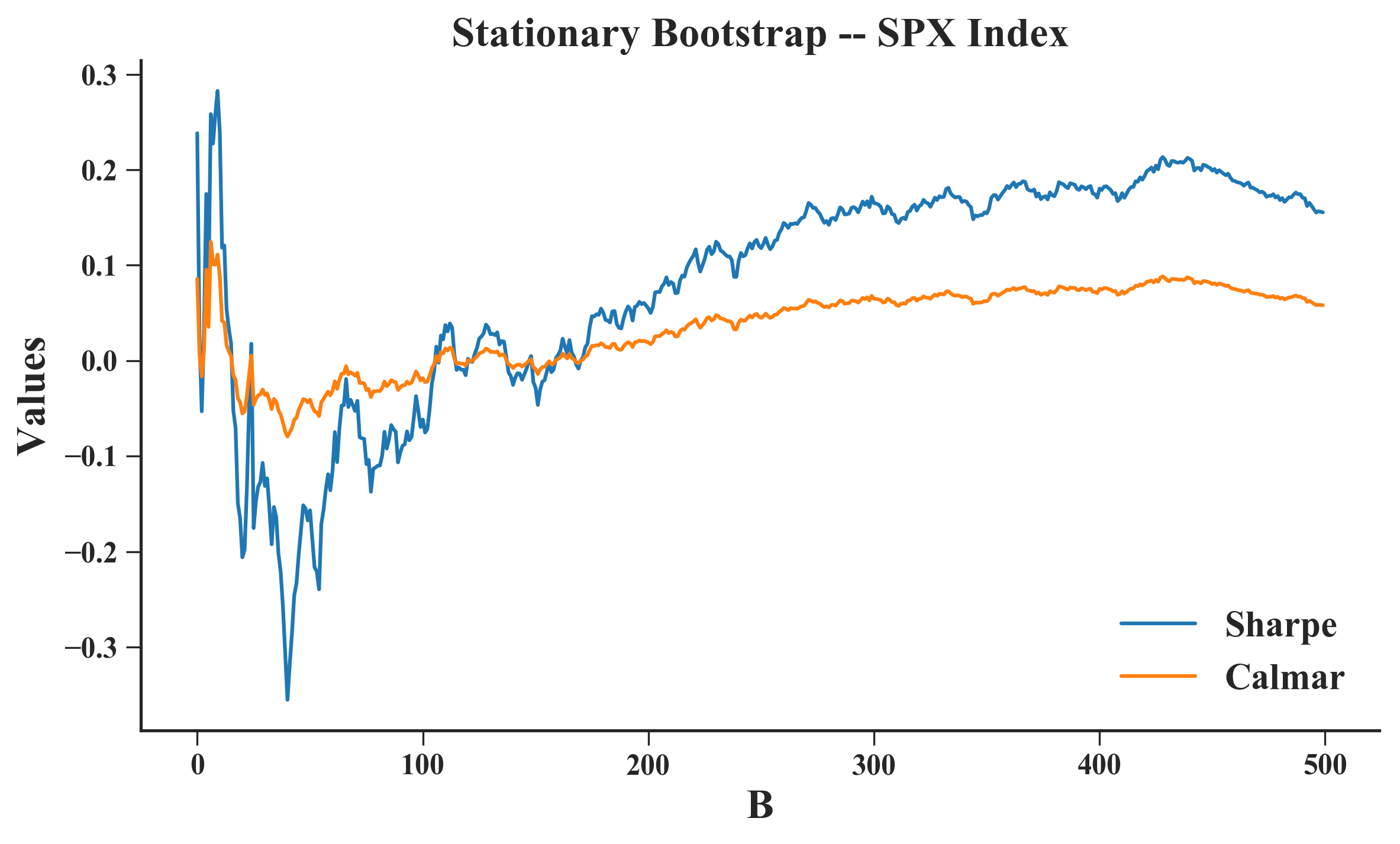}} \\
	\subfloat[]{\includegraphics[width=.85\columnwidth]{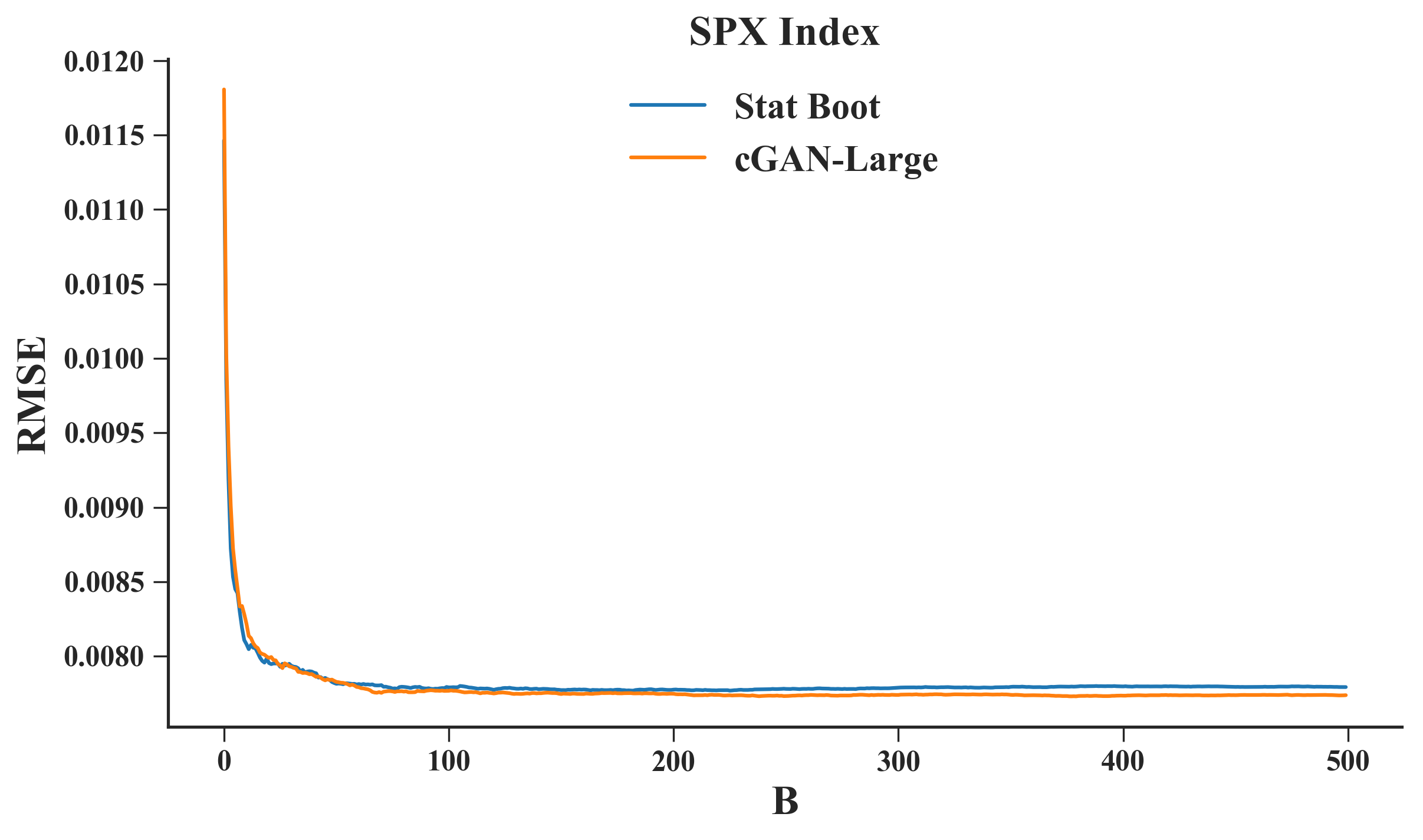}} \subfloat[]{\includegraphics[width=.85\columnwidth]{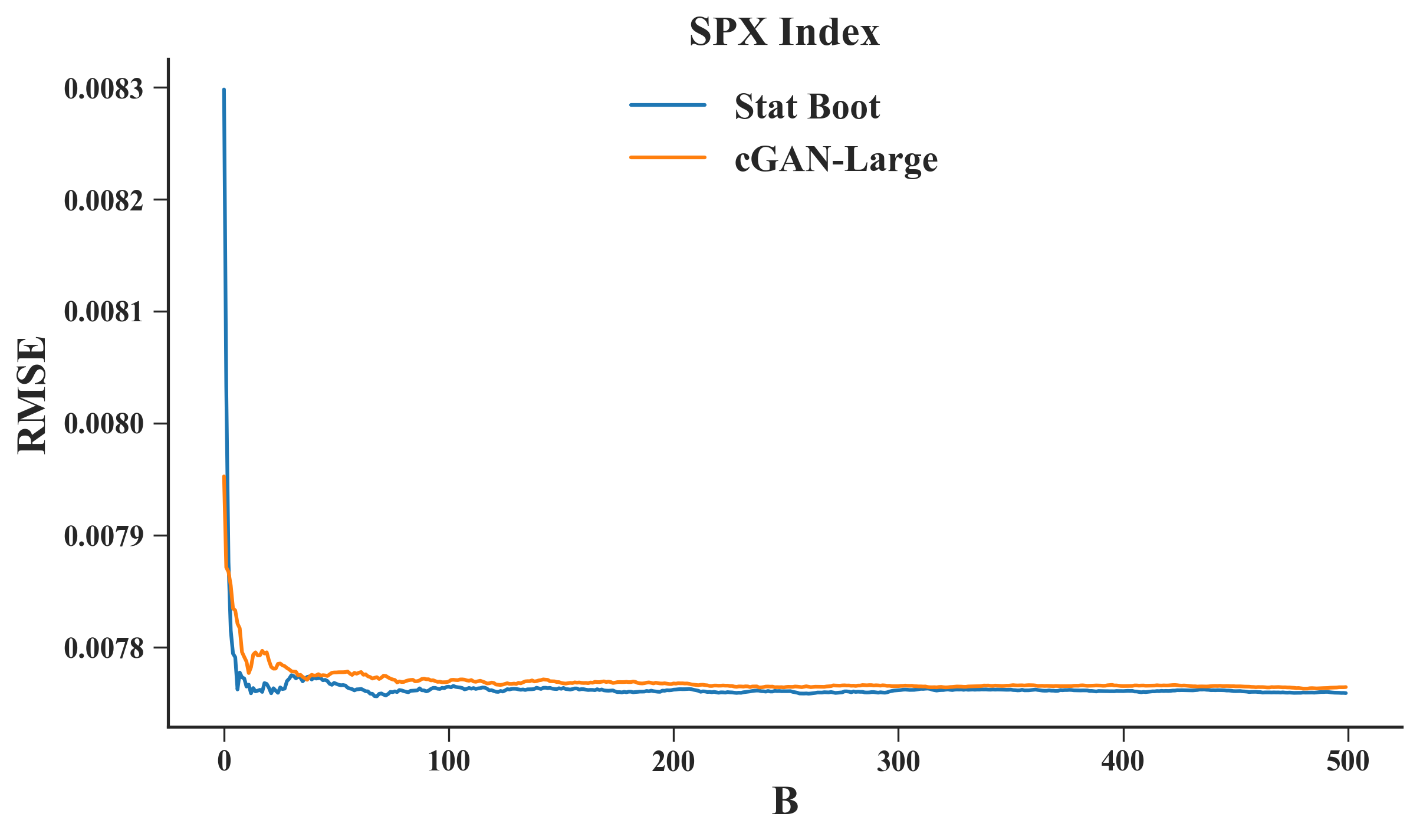}} \\
	\subfloat[]{\includegraphics[width=.85\columnwidth]{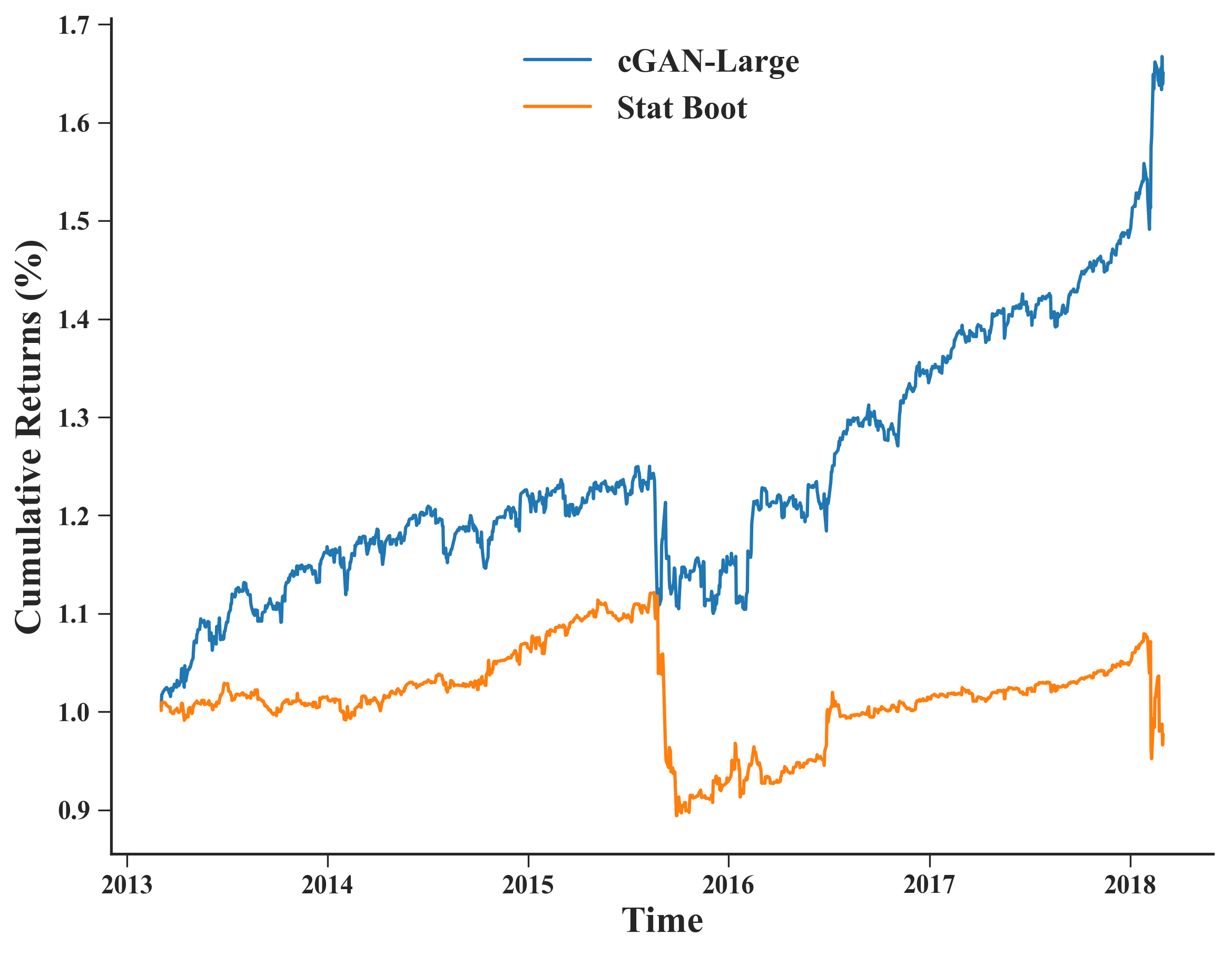}} \subfloat[]{\includegraphics[width=.85\columnwidth]{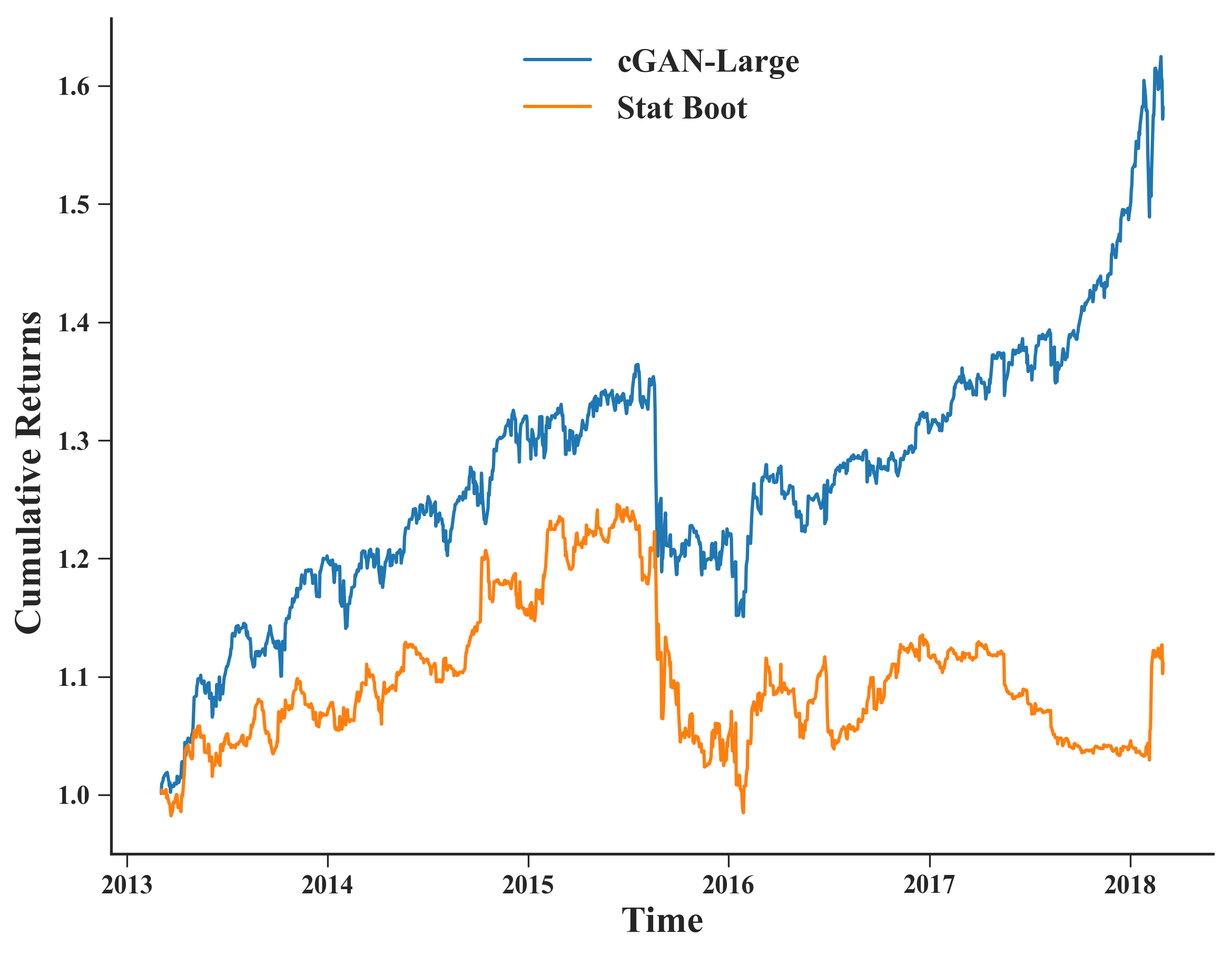}}
	\caption{Main findings for SPX Index; (a-d) Sharpe and Calmar ratios per additional unit in the ensemble; (a,c) Regression Trees built on cGAN-Large and Stat Boot samples, respectively; (b,d) Multilayer Perceptron built on cGAN-Large and Stat Boot samples, respectively. Figures (e,f) outline the RMSE of both approaches per additional unit in the ensemble; (e) Regression Tree, (f) Multilayer Perceptron. Figures (g,h) present the cumulative returns for $B=500$ using (g) Regression Tree and (h) Multilayer Perceptron (targeting 10 \% of volatility per year).}
	\label{spx-combination}
\end{figure*}

To give a more concrete example of this complementarity, Figure \ref{spx-combination} presents the main findings obtained for SPX Index. Figures \ref{spx-combination}a and \ref{spx-combination}b show cGAN-Large as the ensemble strategy using Regression Tree and Multilayer Perceptron as the base learners, respectively. Regression Trees seemed more successful, obtaining a Sharpe and Calmar ratios of 1.00 and 0.75 approximately; but for both methods, cGAN-Large managed to produce positive Sharpe and Calmar ratios. Conversely, Stat Boot failed in both cases, scoring a Sharpe ratio near to 0.0 (Figures \ref{spx-combination}c and \ref{spx-combination}d). This outperformance manifested in a substantial gap between the cumulative returns of the different approaches (Figures \ref{spx-combination}g and \ref{spx-combination}h). Finally, although both methods similarly minimized RMSE (Figures \ref{spx-combination}e and \ref{spx-combination}f), this minimization manifested itself very differently from a Sharpe/Calmar ratio points of view. As a side note, this suggest that minimizing RMSE (a predictive metric) is not an ideal criteria when Sharpe ratio (a financial metric) is the metric that will decide which strategy to be implemented.   

\subsection{Case II: Fine-tuning of Trading Strategies}

Table \ref{main-results-case1-median} presents the quantiles of Sharpe and Calmar ratios in the $OS$ set across the 579 assets for different trading strategies and model validation schemes. Starting from Ridge, we can spot that there not much differences between the model validation schemes, with Naive yielding the worst median (50\%) values (0.121), and hv-Block, Block and cGAN-Medium with the best median (0.138); same can be said with respect to Calmar ratios. 

\begin{table*}[h!]
	\tiny
	\centering
	\caption{Quantiles of Sharpe and Calmar ratios in the $OS$ set across the 579 assets for different trading strategies and model validation schemes.} \label{main-results-case1-median}
	\begin{tabular}{c|c|c|cccccccccc}
		\hline
		\hline
		\multirow{2}{*}{Trad Strat} & \multirow{2}{*}{Metric} & \multirow{2}{*}{Quant} & \multicolumn{10}{c}{Model Validation Scheme} \\
		& & & Naive & One-Split & Sliding & hv-Block & Block & k-Fold & Stat Boot & cGAN-Small & cGAN-Medium & cGAN-Large \\
		\hline
		\multirow{10}{*}{Ridge} & \multirow{5}{*}{Sharpe} & 0\% & -1.594 & -1.594 & -1.594 & -1.594 & -1.594 & -1.493 & -1.594 & -1.493 & -1.493 & -1.493 \\
		& & 25\% & -0.215 & -0.212 & -0.213 & -0.201 & -0.201 & -0.214 & -0.213 & -0.197 & -0.197 & -0.197 \\
		& & 50\% & 0.121 & 0.134 & 0.122 & 0.138 & 0.138 & 0.135 & 0.135 & 0.135 & 0.138 & 0.136 \\
		& & 75\% & 0.403 & 0.418 & 0.409 & 0.409 & 0.409 & 0.424 & 0.410 & 0.424 & 0.419 & 0.419 \\
		& & 100\% & 3.156 & 3.177 & 3.238 & 3.226 & 3.226 & 3.177 & 3.203 & 3.226 & 3.226 & 3.226 \\ \cline{2-13}
		& \multirow{5}{*}{Calmar} & 0\% & -0.290 & -0.290 & -0.218 & -0.290 & -0.290 & -0.290 & -0.209 & -0.218 & -0.203 & -0.203 \\
		& & 25\% & -0.075 & -0.071 & -0.071 & -0.071 & -0.071 & -0.071 & -0.071 & -0.071 & -0.071 & -0.071 \\
		& & 50\% & 0.055 & 0.063 & 0.060 & 0.064 & 0.064 & 0.063 & 0.064 & 0.063 & 0.064 & 0.064 \\
		& & 75\% & 0.236 & 0.251 & 0.232 & 0.239 & 0.241 & 0.241 & 0.241 & 0.241 & 0.244 & 0.244 \\
		& & 100\% & 5.074 & 4.561 & 4.561 & 4.561 & 4.561 & 4.561 & 4.561 & 4.561 & 4.561 & 4.561 \\
		\hline
		\multirow{10}{*}{MLP} & \multirow{5}{*}{Sharpe} & 0\% & -1.362 & -1.583 & -1.554 & -1.291 & -1.297 & -1.389 & -1.062 & -1.150 & -1.212 & -1.176 \\
		& & 25\% & -0.310 & -0.280 & -0.263 & -0.246 & -0.254 & -0.207 & -0.226 & -0.290 & -0.241 & -0.247 \\
		& & 50\% & 0.020 & 0.061 & 0.073 & 0.086 & 0.097 & 0.112 & 0.115 & 0.059 & 0.061 & 0.067 \\
		& & 75\% & 0.352 & 0.390 & 0.400 & 0.396 & 0.416 & 0.406 & 0.429 & 0.380 & 0.396 & 0.416 \\
		& & 100\% & 1.249 & 1.579 & 1.390 & 1.564 & 1.903 & 1.663 & 1.896 & 1.733 & 1.757 & 1.464 \\ \cline{2-13}
		& \multirow{5}{*}{Calmar} & 0\%  & -0.330 & -0.324 & -0.254 & -0.264 & -0.266 & -0.328 & -0.213 & -0.286 & -0.238 & -0.276 \\
		& & 25\% & -0.099 & -0.095 & -0.089 & -0.081 & -0.090 & -0.073 & -0.073 & -0.093 & -0.087 & -0.093 \\
		& & 50\% & 0.008 & 0.026 & 0.031 & 0.039 & 0.046 & 0.050 & 0.056 & 0.022 & 0.028 & 0.032 \\
		& & 75\% & 0.194 & 0.213 & 0.212 & 0.214 & 0.242 & 0.229 & 0.235 & 0.211 & 0.229 & 0.224 \\
		& & 100\% & 1.565 & 1.981 & 1.738 & 2.399 & 2.381 & 1.724 & 1.586 & 2.209 & 1.554 & 1.579 \\
		\hline
		\multirow{10}{*}{GBT} & \multirow{5}{*}{Sharpe} & 0\% & -1.197 & -1.171 & -1.155 & -1.289 & -1.143 & -1.038 & -1.073 & -1.157 & -1.275 & -1.157 \\
		& & 25\% & -0.233 & -0.192 & -0.208 & -0.167 & -0.143 & -0.212 & -0.214 & -0.239 & -0.209 & -0.224 \\
		& & 50\% & 0.088 & 0.159 & 0.142 & 0.175 & 0.211 & 0.174 & 0.162 & 0.123 & 0.150 & 0.133 \\
		& & 75\% & 0.391 & 0.503 & 0.488 & 0.537 & 0.546 & 0.534 & 0.527 & 0.446 & 0.531 & 0.473 \\
		& & 100\% & 5.174 & 5.174 & 4.411 & 5.174 & 5.174 & 5.174 & 5.174 & 3.443 & 1.929 & 5.174 \\ \cline{2-13}
		& \multirow{5}{*}{Calmar} & 0\% & -0.665 & -0.218 & -0.251 & -0.300 & -0.346 & -0.393 & -0.198 & -0.246 & -0.471 & -0.222 \\
		& & 25\% & -0.080 & -0.070 & -0.078 & -0.058 & -0.060 & -0.076 & -0.077 & -0.084 & -0.076 & -0.084 \\
		& & 50\% & 0.038 & 0.071 & 0.066 & 0.081 & 0.105 & 0.077 & 0.072 & 0.053 & 0.067 & 0.064 \\
		& & 75\% & 0.232 & 0.325 & 0.299 & 0.348 & 0.385 & 0.331 & 0.333 & 0.295 & 0.325 & 0.306 \\
		& & 100\% & 7.492 & 7.492 & 6.454 & 7.492 & 7.492 & 7.492 & 7.492 & 3.782 & 2.845 & 7.492 \\
		\hline
		\hline
	\end{tabular}
\end{table*}

Regarding Multilayer Perceptron (MLP) Sharpe ratio results, we can spot a bigger contrast in median terms: Naive fared worst as expected (0.020), with Stationary Bootstrap (Stat Boot) obtaining a median value six fold bigger than Naive. In this case, cGAN-Large (0.067) fared better across the cGANs, but still far from the top median values. For Gradient Boosting Trees (GBT), cGAN-Medium was the best of all cGAN approaches, obtaining better results than Sliding window scheme. However, these figures fell short to those of Block and hv-Block schemes, both faring 0.211 and 0.175 median Sharpe ratios, respectively.

So far we have focused on mainly at the median values, and though we can spot some discrepancies across the methods, these become small when we take into account the average interquartile range\footnote{A measure of dispersion calculated by taking the difference between the 3rd quartile (75\%) and 25\% 1st quartile.} of 0.4 units of Sharpe ratio, around 3-5 times the size of the median values. In this sense, to statistically assess whether some of the observed difference is substantial, Table \ref{RankFriedmanHolmIR} presents a statistical analysis using the average ranks\footnote{When we rank the model validation schemes for a given asset, it means that we sort all them in such way that the best performer is in the first place (receive value equal to 1), the second best is positioned in the second rank (receive value equal to 2), and so on. We can repeat this process for all assets and compute metrics, such as the average rank (e.g., 1.35 means that a particular scheme was placed mostly near to the first place).}, Friedman $\chi^2$ test and Holm correction for multiple hypothesis testing of the different model validation schemes for Ridge, MLP and GBT based on the Sharpe ratio results.

\begin{table*}[h!]
	\centering
	\tiny
	\caption{Average ranks, Friedman and Holm post-hoc statistical tests of the Sharpe ratio for Ridge, MLP and GBT.} \label{RankFriedmanHolmIR}
	\begin{tabular}{ccc|ccc|ccc|c}
		\hline
		\hline
		\multicolumn{3}{c}{Ridge-Sharpe} & \multicolumn{3}{c}{MLP-Sharpe} & \multicolumn{3}{c}{GBT-Sharpe} & \\
		\hline
		Method & Avg Rank & p-value & Method & Avg Rank & p-value & Method & Avg Rank & p-value & Holm Correction \\
		\hline
		\textbf{Naive} & 5.700  & 0.0022 & \textbf{Naive} & 5.900 & $<$ 0.0001 & \textbf{Naive} & 6.074 & $<$ 0.0001 & 0.0055 \\
		Sliding & 5.630 & 0.0081 & \textbf{One-Split} & 5.718  & 0.0004 & \textbf{cGAN-Large} & 5.642  & $<$ 0.0001 & 0.0062 \\
		Stat Boot & 5.605 & 0.0121 & cGAN-Small & 5.549 & 0.0114 & \textbf{Sliding} & 5.628 & $<$ 0.0001 & 0.0071 \\
		k-Fold & 5.592  & 0.0148 & cGAN-Medium & 5.497 & 0.0254 & \textbf{cGAN-Small} & 5.597 & $<$ 0.0001 & 0.0083 \\
		One-Split & 5.510 & 0.0481 & Sliding & 5.489 & 0.0287 & \textbf{cGAN-Medium} & 5.431 & 0.0032 & 0.0100 \\
		cGAN-Small & 5.503  & 0.0531 & hv-Block & 5.449 & 0.0492 & \textbf{k-Fold} & 5.427 & 0.0034 & 0.0125 \\
		cGAN-Large & 5.487  & 0.0644 & Block & 5.444 & 0.0525 & \textbf{Stat Boot} & 5.427 & 0.0034 & 0.0167 \\
		cGAN-Medium & 5.477  & 0.0729 & cGAN-Large & 5.411 & 0.0783 & \textbf{One-Split} & 5.415 & 0.0043 & 0.0250  \\
		hv-Block & 5.253  & 0.4743 & k-Fold & 5.359 & 0.1368 & \textbf{Block} & 5.359 & 0.0112 & 0.0500 \\
		Block & 5.243  & - & Stat Boot & 5.183 & - & hv-Block & 4.992 & - & - \\
		\hline
		Friedman $\chi^2$ & 5715.01 & \textbf{$<$0.0001} & Friedman $\chi^2$ & 2040.8 & \textbf{$<$0.0001} &  Friedman $\chi^2$ & 2865.34 & \textbf{$<$0.0001} &  \\
		\hline
		\hline
	\end{tabular}
\end{table*}

For Ridge Regression, the lowest rank was obtained by Block cross-validation (Block), whilst the worst by Naive (5.700). cGANs methods were consecutively in the third, fourth and fifth places, beating other methods, such as Stat Boot, k-fold cross-validation, etc. The Friedman $\chi^2$ statistics of 5715.01 signal that the hypothesis of equal average rank across the approaches is not statistically credible (p-value $<$ 0.0001). By running a pairwise comparison between Block and the remaining approaches, we can spot that only Naive has stand out as a substantially worst approach, even when we control for multiple hypothesis testing (check Holm Correction column for the adjusted level of significance). 

In respect to MLP ranking results, Naive performed worst as well (5.900), with Stat Boot being the top scheme in this case (5.183); cGAN-Large in the third position, comparing favourably to the other cGAN configurations, as well as hv-Block, Sliding window, etc. Apart from Naive and One-Split/Single holdout scheme, all the remaining approaches were not statistically different from Stat Boot. On the GBT case, we can spot that hv-Block outperformed all approaches, with the cGANs do not delivering reasonable results in this case. 

\begin{figure}[H]
	\centering
	\subfloat[MLP]{\includegraphics[width=\columnwidth]{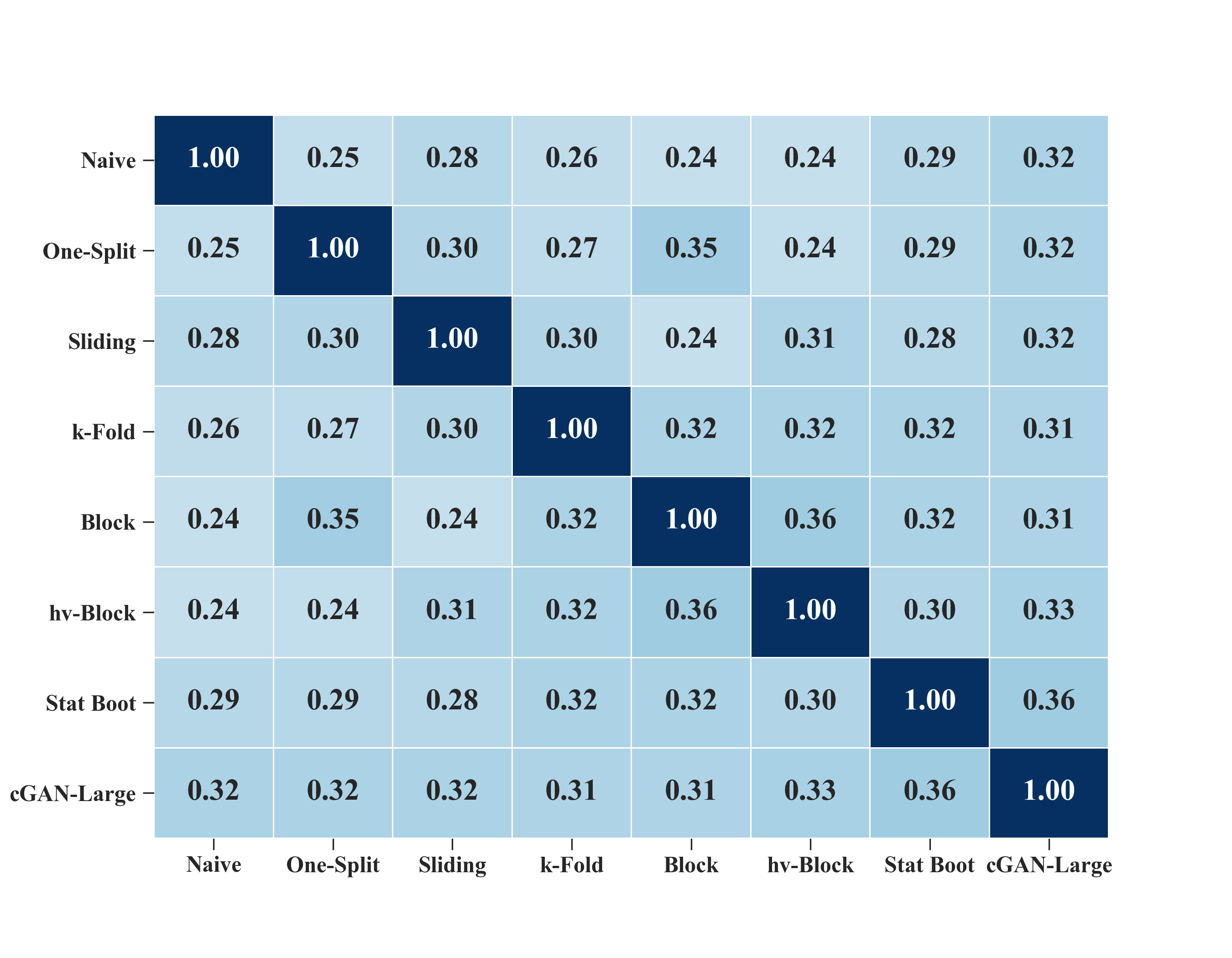}} \\
	\subfloat[GBT]{\includegraphics[width=\columnwidth]{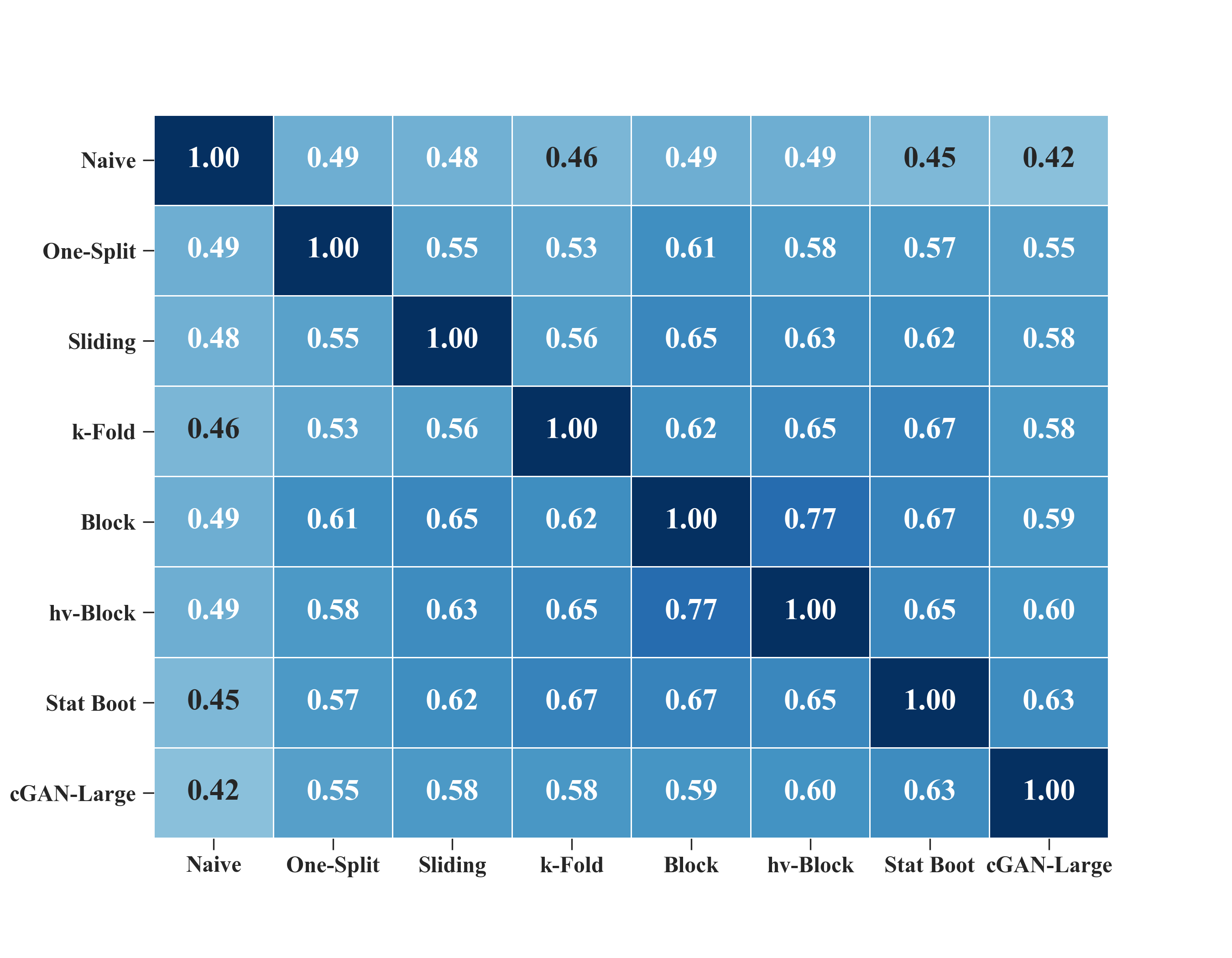}}
	\caption{Correlation matrices based on Sharpe ratios of model validation schemes for MLP (a) and GBT (b).} \label{cgan-corrmatrices}
\end{figure}

Overall, apart from a few analyses and cases (e.g., GBT and Naive method), in aggregate the model validation schemes do not appear to be significantly distinct from each other. This can be interpreted that cGAN is a viable procedure to be part of the fine-tuning pipeline, since its results are statistically indistinguishable to well established methodologies. When we drill down into the results, in particular to the Sharpe ratios of the different approaches, we can spot a low correlation among the validation schemes; Figure \ref{cgan-corrmatrices}\footnote{We decided to omit Ridge since all of the correlations were above 0.8.} presents correlation matrices based on Sharpe ratios of model validation schemes for MLP (a) and GBT (b). 

Though in median and rank terms the strategies look similar, at the micro-level they appear quite the opposite, in particular to the MLP case. Even the cGANs provide distinct Sharpe ratios, showing the importance of the underlying configuration of the Generator/Discriminator. In general, this outline that distinct model validation schemes are arriving with different hyperparameter combinations, incurring in distinct values for Sharpe and Calmar ratios in the $OS$ set. Hence, it may be that in some assets cGAN outcompeted the remaining model validation schemes. To exemplify that, Table \ref{special-cases-mlp} presents a sample of Sharpe ratio results in the $OS$ set for cases where cGAN-Large outcompeted the other methods.

\begin{table}[h!]
	\centering
	\tiny
	\caption{A sample of Sharpe ratio results in the $OS$ set for cases where cGAN-Large outcompeted the other methods.} \label{special-cases-mlp}
	\begin{tabular}{c|ccccc}
		\hline
		\hline
		\multirow{3}{*}{MV Scheme} & ADSWAP2Q  & CADJPY  & ED UN  & NKY  & NZDUSD  \\
		& CMPN  & BGN  & Equity & Index & BGN  \\
		& Curncy & Curncy & & & Curncy \\
		\hline
		Naive & -0.3477 & 0.2009 & -0.0343 & -0.5785 & 0.4503 \\
		One-Split & -0.0184 & -0.4218 & 0.0108 & 0.0520 & -0.0809 \\
		Sliding & 0.5600 & -0.8328 & -0.227 & 0.1083 & 0.1034 \\
		k-Fold & 0.0505 & -0.2861 & 0.4068 & -0.3347 & -0.3104 \\
		Block & 0.4344 & 0.2219 & -0.0971 & -0.8870 & 0.1215 \\
		hv-Block & 0.1120 & -0.3932 & 0.5364 & -0.0244 & -0.272 \\
		Stat Boot & 0.4296 & -0.19498 & 0.3107 & -0.3068 & -0.2616 \\
		cGAN-Small & 0.5146 & -0.6222 & -0.0980 & 0.1095 & 0.3059 \\
		cGAN-Medium & 0.84 & -0.0901 & 0.3443 & 0.0884 & 0.0582 \\
		\textbf{cGAN-Large} & 1.4207 & 0.5885 & 1.1224 & 0.2263 & 0.6703 \\
		\hline
		\hline
	\end{tabular}
\end{table}

We can spot a few instances that cGAN-Large substantially fared better results, such as in a 2y Australian Dollar Swap, New Zealand Dollar vs US Dollar Currency, and Consolidated Edison Inc. Equity. This set of results suggest that cGAN-Large is a viable alternative for fine-tuning machine learning models when other methodologies provide poor results, and it should be considered in the portfolio of different validation schemes aside of the distinct trading strategies models.

\begin{figure}[h!]
	\centering
	\includegraphics[width=\columnwidth]{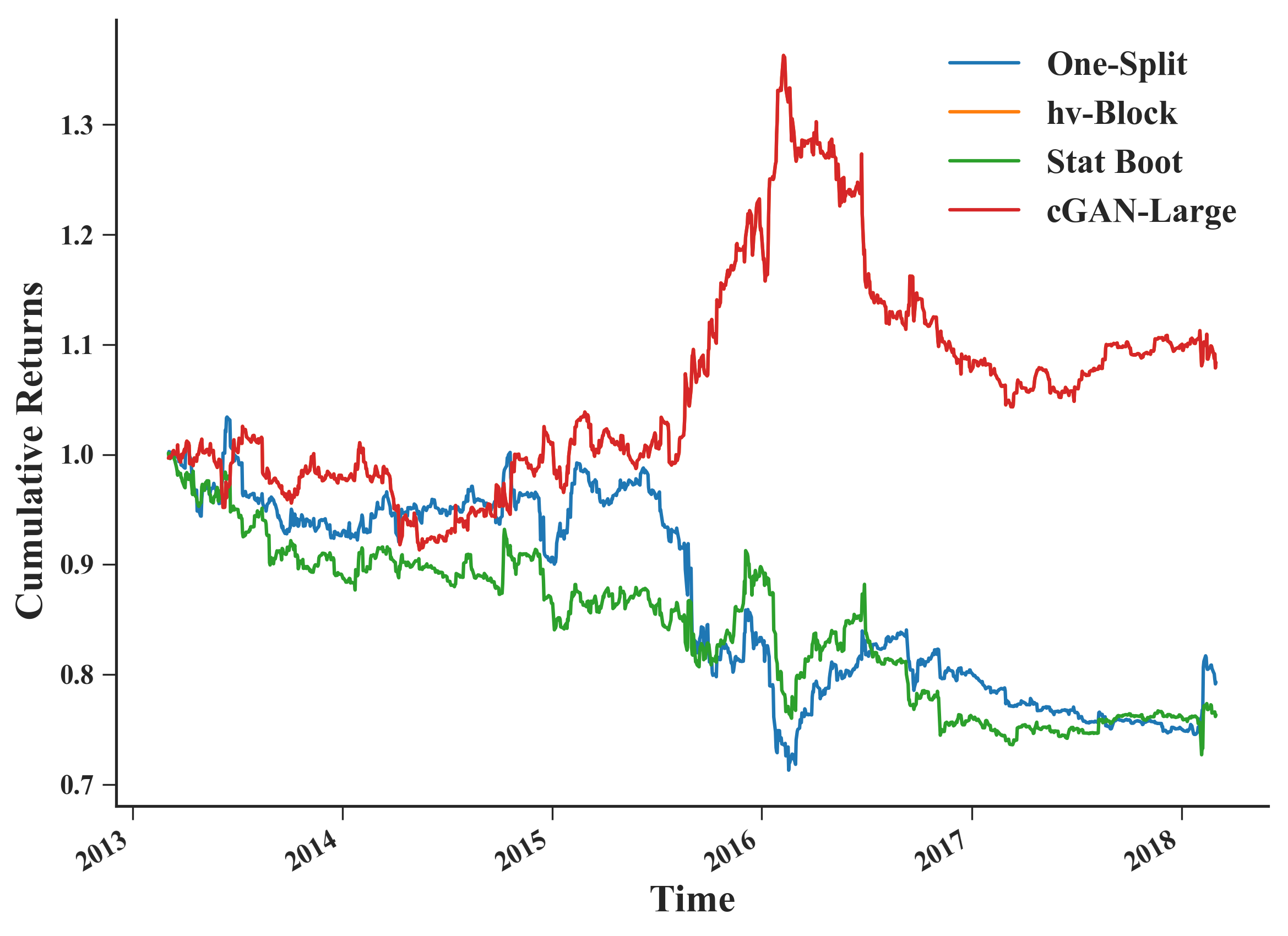}
	\caption{SPX Index cumulative returns in the $OS$ set for different model validation schemes using MLP as the trading strategy. cGAN-Large and hv-Block found out the same hyperparameters, therefore obtaining similar profiles.} \label{spx-mlp-mvschemes}
\end{figure}

Finally, Figure \ref{spx-mlp-mvschemes} outlines the cumulative returns for SPX Index for a few of the different model validation schemes using MLP as the trading strategy. In this case, Stat Boot and One-Split were unable to produce a profit after five years of trading, whilst cGAN-Large and hv-Block produced around 10\% of return for a given initial amount of investment (they both found out the same hyperparameters, therefore obtaining similar profiles). This is another example that demonstrate the relevance of having a set of model assessment schemes, as similar as the more common defensive posture of having a portfolio of trading strategies/models and hyperparameter optimization schemes. 



\section{Conclusion}

This work has proposed the use of Conditional Generative Adversarial Networks (cGANs) for trading strategies calibration and aggregation. This emerging technique can have an impact into aspects of trading strategies, specifically fine-tuning and to form ensembles. Also, we can list a few advantages of such method, like: (i) generating more diverse training and testing sets, compared to traditional resampling techniques; (ii) able to draw samples specifically about stressful events, ideal for model checking and stress testing; and (iii) providing a level of anonymization to the dataset, differently from other techniques that (re)shuffle/resample data. 

The price paid is having to fit this generative model for a given time series. To this purpose, we provided a full methodology on: (i) the training and selection of a cGAN for time series generation; (ii) how each sample is used for strategies calibration; and (iii) how all generated samples can be used for ensemble modelling. To provide evidence that our approach is well grounded, we have designed an experiment encompassing 579 assets, tested multiple trading strategies, and analysed different capacities for Generator/Discriminator. In summary, our main contributions were to show that our procedure to train and select cGANs is sound, as well as able to obtain competitive results against traditional methods for fine-tuning and ensemble modelling.

Being more specific, in the Case Study I: Combination of Trading Strategies, we compared cGAN Sampling and Aggregation with Stationary Bootstrap. Our results suggest that both approaches are equivalent in aggregate, with non-statistically significant advantage for cGAN when using Regression Trees, and for Stationary Bootstrap when using a shallow Multilayer Perceptron. But when Bagging via Stationary Bootstrap fails to perform properly, it is possible to use cGAN Sampling and Aggregation as a tool to combine weak signals in alpha generating strategies; SPX Index was an example where cGAN outcompeted Stationary Bootstrap by a wide margin. 

In relation to the Case Study II: Fine-tuning of trading Strategies, we compared cGAN with a wide range of model validation strategies. All of these were techniques designed to handle time series scenarios: these ranged from window-based methods, to shuffling and reshuffling of a time series. We have evidence that cGANs can be used for model tuning, bearing better results in cases where traditional schemes fail. A side outcome of our work is the wealth of results and comparisons: to the best of our knowledge most of the applied model validation strategies have not yet been cross compared using real datasets and different models.

Finally, our work also open new avenues to future investigations. We list  a few potential extensions and directions for further research:
\begin{itemize}
	\item cGANs for stress testing: a stress test examines the potential impact of a hypothetical adverse scenario on the health of a financial institution, or even a trading strategy. In doing so, stress tests allow the quantitative strategist to assess a strategy resilience to a range of adverse shocks and ensure they are sufficiently hedged to withstand those shocks. A proper benchmark can be model-based bootstrap, since it allows conditional variables which facilitates the process of generating resamples of crisis events.
	
	\item Selection metrics for cGANs: we have adopted the Root Mean Square Error as the loss function between the generator samples and the actual data, however nothing limits the user to use another type of loss function. It could be a metric that take into account several moments and cross-moments of a time series. With financial time series, taking into account stylized facts \cite{cont2001empirical} can be a feasible alternative to produce samples that are more meaningful and resemble more a financial asset return.
	
	\item Combining cGAN with Stationary Bootstrap: in our results section, in particular to Figure \ref{cgan-statboot-metrics}, we observed the low correlation between the Sharpe ratios obtained in both approaches. This imply that a mixed approach, that is, combining resamples from cGAN and Stationary Bootstrap, can yield better results than opting for a single approach. 
	
	\item Extensions and other applications: a natural extension is to consider predicting directly multiple steps ahead, or considering modelling multiple financial time series. Both can improve our results, as well as, may reduce the time to train and select a cGAN. Another  extension is to consider other architectures, such as AdaGANs \cite{tolstikhin2017adagan} or Ensemble of GANs \cite{wang2016ensembles}. Also, other applications such as fine-tuning and combination of time series forecasting methods; a good benchmark are the M3 and M4 competitions \cite{makridakis2000m3,makridakis2018m4} that involve a large number of time series as well as results from a wide array of forecasting methods.
	
\end{itemize}

\section*{Acknowledgment}

The authors would like to thank EO for his insightful suggestions and critical comments about our work. Adriano Koshiyama would like to acknowledge the National Research Council of Brazil for his PhD scholarship, and The Alan Turing Institute for providing infra-structure and environment to conclude this work.

\newpage
\balance
\bibliographystyle{plain}
\bibliography{cganbib}

\begin{thebibliography}{10}

\bibitem{acar2002advanced}
Emmanuel Acar and Stephen Satchell.
\newblock {\em Advanced trading rules}.
\newblock Butterworth-Heinemann, 2002.

\bibitem{arjovsky2017wasserstein}
Martin Arjovsky, Soumith Chintala, and L{\'e}on Bottou.
\newblock Wasserstein generative adversarial networks.
\newblock In {\em International Conference on Machine Learning}, pages
  214--223, 2017.

\bibitem{arlot2010survey}
Sylvain Arlot, Alain Celisse, et~al.
\newblock A survey of cross-validation procedures for model selection.
\newblock {\em Statistics surveys}, 4:40--79, 2010.

\bibitem{bailey2015probability}
David~H Bailey, Jonathan Borwein, Marcos Lopez~de Prado, and Qiji~Jim Zhu.
\newblock The probability of backtest overfitting.
\newblock {\em Journal of Computational Finance (Risk Journals)}, 2015.

\bibitem{bates1969combination}
John~M Bates and Clive~WJ Granger.
\newblock The combination of forecasts.
\newblock {\em Journal of the Operational Research Society}, 20(4):451--468,
  1969.

\bibitem{bergmeir2018note}
Christoph Bergmeir, Rob~J Hyndman, and Bonsoo Koo.
\newblock A note on the validity of cross-validation for evaluating
  autoregressive time series prediction.
\newblock {\em Computational Statistics \& Data Analysis}, 120:70--83, 2018.

\bibitem{bergstra2012random}
James Bergstra and Yoshua Bengio.
\newblock Random search for hyper-parameter optimization.
\newblock {\em Journal of Machine Learning Research}, 13(Feb):281--305, 2012.

\bibitem{cont2001empirical}
Rama Cont.
\newblock Empirical properties of asset returns: stylized facts and statistical
  issues.
\newblock 2001.

\bibitem{creswell2018generative:paper2}
Antonia Creswell, Tom White, Vincent Dumoulin, Kai Arulkumaran, Biswa Sengupta,
  and Anil~A Bharath.
\newblock Generative adversarial networks: An overview.
\newblock {\em IEEE Signal Processing Magazine}, 35(1):53--65, 2018.

\bibitem{derrac2011practical}
Joaqu{\'\i}n Derrac, Salvador Garc{\'\i}a, Daniel Molina, and Francisco
  Herrera.
\newblock A practical tutorial on the use of nonparametric statistical tests as
  a methodology for comparing evolutionary and swarm intelligence algorithms.
\newblock {\em Swarm and Evolutionary Computation}, 1(1):3--18, 2011.

\bibitem{douzas2018effective}
Georgios Douzas and Fernando Bacao.
\newblock Effective data generation for imbalanced learning using conditional
  generative adversarial networks.
\newblock {\em Expert Systems with applications}, 91:464--471, 2018.

\bibitem{efron2016computer}
Bradley Efron and Trevor Hastie.
\newblock {\em Computer age statistical inference}, volume~5.
\newblock Cambridge University Press, 2016.

\bibitem{eggensperger2013towards}
Katharina Eggensperger, Matthias Feurer, Frank Hutter, James Bergstra, Jasper
  Snoek, Holger Hoos, and Kevin Leyton-Brown.
\newblock Towards an empirical foundation for assessing bayesian optimization
  of hyperparameters.
\newblock In {\em NIPS workshop on Bayesian Optimization in Theory and
  Practice}, volume~10, page~3, 2013.

\bibitem{eling2007does}
Martin Eling and Frank Schuhmacher.
\newblock Does the choice of performance measure influence the evaluation of
  hedge funds?
\newblock {\em Journal of Banking \& Finance}, 31(9):2632--2647, 2007.

\bibitem{esteban2017real}
Crist{\'o}bal Esteban, Stephanie~L Hyland, and Gunnar R{\"a}tsch.
\newblock Real-valued (medical) time series generation with recurrent
  conditional gans.
\newblock {\em arXiv preprint arXiv:1706.02633}, 2017.

\bibitem{fiore2017using}
Ugo Fiore, Alfredo De~Santis, Francesca Perla, Paolo Zanetti, and Francesco
  Palmieri.
\newblock Using generative adversarial networks for improving classification
  effectiveness in credit card fraud detection.
\newblock {\em Information Sciences}, 2017.

\bibitem{friedman2001elements}
Jerome Friedman, Trevor Hastie, and Robert Tibshirani.
\newblock {\em The elements of statistical learning}, volume~1.
\newblock Springer series in statistics New York, NY, USA:, 2001.

\bibitem{goodfellow2014generative}
Ian Goodfellow, Jean Pouget-Abadie, Mehdi Mirza, Bing Xu, David Warde-Farley,
  Sherjil Ozair, Aaron Courville, and Yoshua Bengio.
\newblock Generative adversarial nets.
\newblock In {\em Advances in neural information processing systems}, pages
  2672--2680, 2014.

\bibitem{gulrajani2017improved}
Ishaan Gulrajani, Faruk Ahmed, Martin Arjovsky, Vincent Dumoulin, and Aaron~C
  Courville.
\newblock Improved training of wasserstein gans.
\newblock In {\em Advances in Neural Information Processing Systems}, pages
  5767--5777, 2017.

\bibitem{harvey2015backtesting}
Campbell~R Harvey and Yan Liu.
\newblock Backtesting.
\newblock {\em The Journal of Portfolio Management}, pages 12--28, 2015.

\bibitem{hsiao2014there}
Cheng Hsiao and Shui~Ki Wan.
\newblock Is there an optimal forecast combination?
\newblock {\em Journal of Econometrics}, 178:294--309, 2014.

\bibitem{huang2017stacked}
Xun Huang, Yixuan Li, Omid Poursaeed, John~E Hopcroft, and Serge~J Belongie.
\newblock Stacked generative adversarial networks.
\newblock In {\em CVPR}, volume~2, page~3, 2017.

\bibitem{jiang2017markov}
Gaoxia Jiang and Wenjian Wang.
\newblock Markov cross-validation for time series model evaluations.
\newblock {\em Information Sciences}, 375:219--233, 2017.

\bibitem{lahiri2013resampling}
Soumendra~Nath Lahiri.
\newblock {\em Resampling methods for dependent data}.
\newblock Springer Science \& Business Media, 2013.

\bibitem{loshchilov2016cma}
Ilya Loshchilov and Frank Hutter.
\newblock Cma-es for hyperparameter optimization of deep neural networks.
\newblock {\em arXiv preprint arXiv:1604.07269}, 2016.

\bibitem{makridakis2000m3}
Spyros Makridakis and Michele Hibon.
\newblock The m3-competition: results, conclusions and implications.
\newblock {\em International journal of forecasting}, 16(4):451--476, 2000.

\bibitem{makridakis2018m4}
Spyros Makridakis, Evangelos Spiliotis, and Vassilios Assimakopoulos.
\newblock The m4 competition: Results, findings, conclusion and way forward.
\newblock {\em International Journal of Forecasting}, 2018.

\bibitem{mao2017least}
Xudong Mao, Qing Li, Haoran Xie, Raymond~YK Lau, Zhen Wang, and Stephen~Paul
  Smolley.
\newblock Least squares generative adversarial networks.
\newblock In {\em Computer Vision (ICCV), 2017 IEEE International Conference
  on}, pages 2813--2821. IEEE, 2017.

\bibitem{mirza2014conditional}
Mehdi Mirza and Simon Osindero.
\newblock Conditional generative adversarial networks.
\newblock {\em Manuscript: https://arxiv. org/abs/1709.02023}, 2014.

\bibitem{ostrovski2018autoregressive}
Georg Ostrovski, Will Dabney, and R{\'e}mi Munos.
\newblock Autoregressive quantile networks for generative modeling.
\newblock {\em arXiv preprint arXiv:1806.05575}, 2018.

\bibitem{racine2000consistent}
Jeff Racine.
\newblock Consistent cross-validatory model-selection for dependent data:
  hv-block cross-validation.
\newblock {\em Journal of econometrics}, 99(1):39--61, 2000.

\bibitem{radford2015unsupervised}
Alec Radford, Luke Metz, and Soumith Chintala.
\newblock Unsupervised representation learning with deep convolutional
  generative adversarial networks.
\newblock {\em arXiv preprint arXiv:1511.06434}, 2015.

\bibitem{romano2016efficient}
Joseph~P Romano and Michael Wolf.
\newblock Efficient computation of adjusted p-values for resampling-based
  stepdown multiple testing.
\newblock {\em Statistics \& Probability Letters}, 113:38--40, 2016.

\bibitem{salimans2016improved}
Tim Salimans, Ian Goodfellow, Wojciech Zaremba, Vicki Cheung, Alec Radford, and
  Xi~Chen.
\newblock Improved techniques for training gans.
\newblock In {\em Advances in Neural Information Processing Systems}, pages
  2234--2242, 2016.

\bibitem{sharpe1994sharpe}
William~F Sharpe.
\newblock The sharpe ratio.
\newblock {\em Journal of portfolio management}, 21(1):49--58, 1994.

\bibitem{timmermann2006forecast}
Allan Timmermann.
\newblock Forecast combinations.
\newblock {\em Handbook of economic forecasting}, 1:135--196, 2006.

\bibitem{tolstikhin2017adagan}
Ilya~O Tolstikhin, Sylvain Gelly, Olivier Bousquet, Carl-Johann Simon-Gabriel,
  and Bernhard Sch{\"o}lkopf.
\newblock Adagan: Boosting generative models.
\newblock In {\em Advances in Neural Information Processing Systems}, pages
  5424--5433, 2017.

\bibitem{wang2018evolutionary}
Chaoyue Wang, Chang Xu, Xin Yao, and Dacheng Tao.
\newblock Evolutionary generative adversarial networks.
\newblock {\em arXiv preprint arXiv:1803.00657}, 2018.

\bibitem{wang2016ensembles}
Yaxing Wang, Lichao Zhang, and Joost van~de Weijer.
\newblock Ensembles of generative adversarial networks.
\newblock {\em arXiv preprint arXiv:1612.00991}, 2016.

\bibitem{young1991calmar}
Terry~W Young.
\newblock Calmar ratio: A smoother tool.
\newblock {\em Futures}, 20(1):40, 1991.

\bibitem{zhao2016energy}
Junbo Zhao, Michael Mathieu, and Yann LeCun.
\newblock Energy-based generative adversarial network.
\newblock {\em arXiv preprint arXiv:1609.03126}, 2016.

\bibitem{zhou2018stock}
Xingyu Zhou, Zhisong Pan, Guyu Hu, Siqi Tang, and Cheng Zhao.
\newblock Stock market prediction on high-frequency data using generative
  adversarial nets.
\newblock {\em Mathematical Problems in Engineering}, 2018, 2018.

\end{thebibliography}

\end{document}